\documentclass{article} % For LaTeX2e
\usepackage{style,times}

\usepackage[utf8]{inputenc} % allow utf-8 input
\usepackage[T1]{fontenc}    % use 8-bit T1 fonts
\usepackage{hyperref}       % hyperlinks
\usepackage{url}            % simple URL typesetting
\usepackage{booktabs}       % professional-quality tables
\usepackage{amsfonts}       % blackboard math symbols
\usepackage{nicefrac}       % compact symbols for 1/2, etc.
\usepackage{microtype}      % microtypography
\usepackage{xcolor, colortbl}         % colors
\usepackage{graphicx}

\usepackage{multirow}
\usepackage{amsmath}
\usepackage{amssymb}
\usepackage{algorithm}
\usepackage{algorithmic}
\usepackage{makecell}
\usepackage{enumitem}
\usepackage{caption}
\usepackage{wrapfig}
\usepackage{pifont}
\newcommand{\cmark}{\ding{51}}%
\newcommand{\xmark}{\ding{55}}%

\definecolor{citeblue}{RGB}{0,130,180}
\definecolor{refrose}{RGB}{199,21,133}
\hypersetup{
    colorlinks=true,
    citecolor=citeblue,
    linkcolor=refrose,
    urlcolor=citeblue
}

\title{Do Better Visual Representations Always Lead to Better End-to-End Autonomous Driving?}

\author{%
  Zihao Zhang$^{1,2,3}$\quad
  Haochen Tian$^{2}$\quad
  Tianyu Li$^{2}$\quad
  Changhui Jing$^{2}$\quad
  Jingliang He$^{3}$\\
  \textbf{Naisheng Ye}$^{2}$\quad
  \textbf{Ziyuan Pu}$^{1 \ddagger}$\quad
  \textbf{Zhenjie Yang}$^{2\ddagger}$\\
  \\
  $^{1}$Southeast University \quad $^{2}$OpenDriveLab at The University of Hong Kong \\
  $^{3}$Shenzhen Loop Area Institute \\
  $^{\ddagger}$Correspondence Authors.
  ~~Primary contact:
  {\hypersetup{hidelinks}\href{mailto:zihao.zhang@opendrivelab.com} {\texttt{zihao.zhang@opendrivelab.com}}}
}

\iclrfinalcopy
\begin{document}

\maketitle
\fancyhead{}
\lhead{Preprint}
\renewcommand{\headrulewidth}{0.4pt}

\begin{abstract}
Visual foundation models (VFMs) are increasingly integrated into end-to-end autonomous driving for their powerful representations, yet it remains unclear when these representations improve driving performance.
To investigate this question, we introduce ViRA, a planner-agnostic visual representation alignment framework that keeps the planner architecture and inference cost unchanged.
Our study reveals three findings:
(1) VFM-guided visual representations consistently improve driving performance across diverse end-to-end planners, with gains extending to zero-shot closed-loop evaluation.
(2) The choice of VFM target matters for planning performance, and alignment to a different VFM can further benefit planners with pre-trained VFM encoders.
(3) Auxiliary perception supervision reduces sensitivity to VFM target selection, narrowing the EPDMS spread across five targets from 2.7 to 0.5 points and potentially compensating for less effective VFM targets.
Guided by these findings, we develop ViRA-Diffusion, a diffusion-based planner trained without auxiliary perception supervision, which achieves 92.3 EPDMS on NAVSIM v2 navtest, outperforming recent methods in our comparison by at least 1.9 points.
The results motivate jointly considering target selection and planner supervision when integrating VFMs into end-to-end autonomous driving.
The results and demo are available at {\urlstyle{same}\url{https://github.com/OpenDriveLab/ViRA}}.

\end{abstract}
\begin{figure}[!htbp]
\centering
\includegraphics[width=1.0\textwidth]{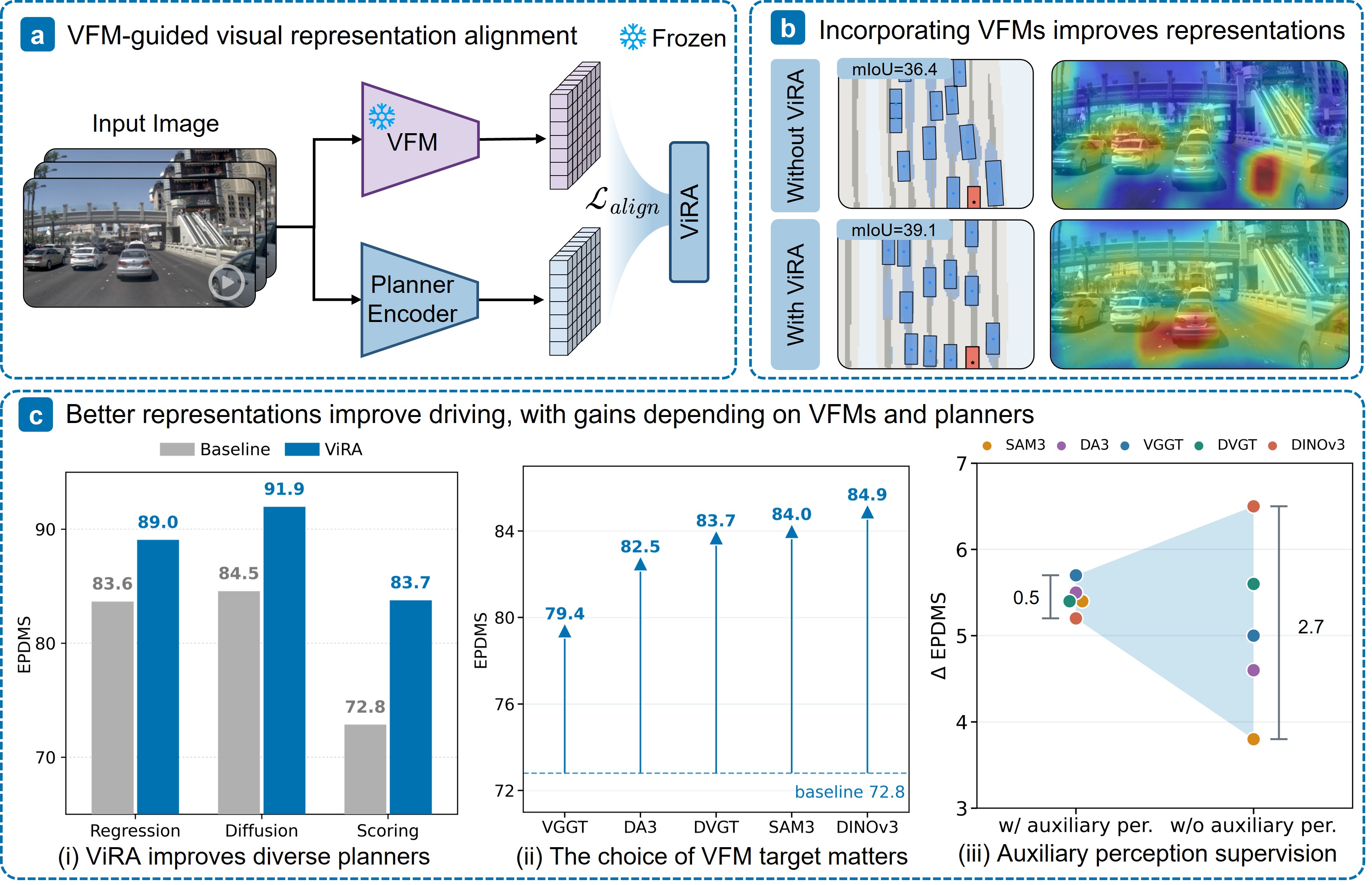}
\caption{\textbf{Better visual representations improve planning, with gains depending on both VFMs and planners.}
(a) ViRA aligns planner representations with a frozen VFM during training, leaving inference unchanged.
(b) Segmentation and visual attention suggest improved representations.
(c) Better visual representations improve planning, but the gains depend on the VFM target and planner.
}
\label{fig:teaser_overview}
\end{figure}
%%%%%%

\section{Introduction}
\label{sec:intro}

End-to-end autonomous driving is considered a promising route to scalable real-world deployment~\citep{chen2024end-to-end, guan2026planning}.
Within this pipeline, visual representations provide the planner with information about the environment, making the learning of high-quality visual representations an important research problem~\citep{jiang2023vad,li2024bevformer}.
As visual foundation models (VFMs) advance in semantic understanding and 3D geometry~\citep{simeoni2025dinov3,wang2025vggt,carion2025sam,lin2025da3}, an increasing number of studies either use VFMs directly~\citep{sima2024drivelm,Wang2026vggdrive,naeinian2026zero} or incorporate their pre-trained representations through distillation and representation alignment~\citep{strong2026learning,li2025sf,xie2026latentvla}.
However, most studies focus on a single planner–VFM pairing~\citep{zuo2026dvgt2,wang2026latent-wam}, and systematic comparisons of planning gains across different pairings remain limited.
It remains unclear whether the advantages of VFMs on visual tasks consistently translate into improved driving performance.

To answer this question, we introduce ViRA, a VFM-guided representation alignment framework illustrated in Figure~\ref{fig:teaser_overview}(a). 
ViRA aligns the planner's intermediate features with those of a frozen VFM target while keeping the planner architecture and inference cost unchanged~\citep{yu2024repa, singh2025irepa,li2025sf}.
Using this framework, we compare different combinations of end-to-end planners and VFM targets on NAVSIM v2~\citep{cao2025navsimv2} and HUGSIM~\citep{zhou2025hugsim}.
We find that VFM-guided visual representations consistently improve driving performance across the diverse planners, but the gains depend on both the VFM target and planner design.
We further examine these findings below, focusing on overall driving gains, the effects of VFM target selection, and the role of auxiliary perception supervision.

\textbf{VFM-guided visual representations improve end-to-end driving performance.}
As shown in Figure~\ref{fig:teaser_overview}(c-i), ViRA improves driving performance across regression-based~\citep{Kashyap2023Transfuser}, diffusion-based~\citep{liao2025diffusiondrive}, and scoring-based~\citep{feng2025rap3drasterizationaugmented} planners, yielding average gains of 7.9 and 8.6 EPDMS points over their respective baselines on NAVSIM v2 navtest and navhard, respectively.
These benefits extend to zero-shot closed-loop evaluation on HUGSIM, where overall HD-Score improves by an average of 4.0 points.
Figure~\ref{fig:teaser_overview}(b) shows that driving gains are accompanied by improved perception performance.
To test whether the gains come from pre-training rather than loss regularization alone, we compare pre-trained and randomly initialized targets under the same alignment objectives.
Randomly initialized targets fail to reproduce these perception and planning gains, supporting the importance of pre-trained VFM representations.
This raises the question of whether VFMs with different pre-training objectives yield different planning benefits.

\textbf{The choice of VFM target matters.}
As shown in Figure~\ref{fig:teaser_overview}(c-ii), gains for the same Rap$^\ast$~\citep{feng2025rap3drasterizationaugmented} architecture range from 6.6 to 12.1 EPDMS points across VFM targets (a 5.5-point spread), with DINOv3~\citep{simeoni2025dinov3} yielding the largest gain.
Yet even when Rap$^\ast$ uses DINOv3 as its pre-trained backbone, alignment to VGGT~\citep{wang2025vggt}, a 3D geometric foundation model, further improves EPDMS from 83.8 to 86.3 (+2.5 points).
This dependence on the VFM target also appears in TransFuser~\citep{Kashyap2023Transfuser} when trained without auxiliary perception supervision, with DINOv3 again performing best.
With auxiliary perception supervision, however, TransFuser and DiffusionDrive~\citep{liao2025diffusiondrive} achieve similar planning performance across these targets.
We hypothesize that auxiliary perception supervision affects how planners learn from VFM targets.

\textbf{Auxiliary perception supervision reduces sensitivity to VFM target selection.}
To examine this factor, we compare TransFuser with and without auxiliary perception supervision.
Across the same five VFM targets, planning gains over the respective baselines without representation alignment range from 5.2 to 5.7 EPDMS points with auxiliary supervision and from 3.8 to 6.5 points without it.
As shown in Figure~\ref{fig:teaser_overview}(c-iii), removing auxiliary supervision widens the spread in gains across targets from 0.5 to 2.7 EPDMS points.
More specifically, the highest score stays nearly unchanged, while the lowest drops from 88.8 to 86.4, indicating that auxiliary supervision mainly compensates for the less effective targets.
These results show that, in this setting, auxiliary perception supervision narrows the differences in gains across alignment targets, making planning performance less sensitive to VFM target selection.

These findings make a concrete prediction: a planner can achieve strong planning performance when its auxiliary perception supervision is removed, and a suitable VFM target is supplied in its place.
We verify this with \textbf{ViRA-Diffusion}, a diffusion-based~\citep{liao2025diffusiondrive} planner trained with ViRA using DINOv3~\citep{simeoni2025dinov3} as the target and no auxiliary perception supervision.
It reaches 92.3 EPDMS on NAVSIM v2~\citep{cao2025navsimv2} navtest, outperforming the recent state-of-the-art method by 1.9 points. We highlight the main contributions of this paper below:

\begin{itemize}[leftmargin=*,topsep=2pt,itemsep=1pt,parsep=0pt,partopsep=0pt]
    \item We systematically study the driving benefits of pre-trained VFM representations across planners and targets using ViRA, a planner-agnostic experimental framework that preserves the deployed planner architecture and inference cost.

    \item We show that VFM target selection affects planning performance, while auxiliary perception supervision narrows performance differences across targets, highlighting the importance of considering target selection and planner supervision jointly.

    \item We demonstrate the practical applicability of VFM-guided alignment through \textbf{ViRA-Diffusion}, which uses DINOv3 alignment to achieve 92.3 EPDMS on NAVSIM v2 navtest without auxiliary perception supervision.
\end{itemize}

\section{Preliminaries}
\label{sec:preli}
\subsection{VFM-Guided End-to-End Planner}

\paragraph{End-to-end Planner.}
Given a single-frame multi-view observation $\mathcal{O}=\{\mathbf{I}^{i}\}_{i=1}^{N}$, an end-to-end planner predicts a future ego trajectory $\hat{\tau}=\{\hat{\mathbf{p}}_k\}_{k=1}^{K}$, where $N$ is the number of camera views and $\hat{\mathbf{p}}_k=(\hat{x}_k,\hat{y}_k)$ denotes the predicted position at future step $k$ in the ego coordinate system.
We formulate the planner as an encoder $E_{\theta}$ followed by a policy head $\pi_{\psi}$:
\begin{equation}
\label{eq:planner_prelim}
\mathbf{F}^{s}=E_{\theta}(\mathcal{O}),
\qquad
\hat{\tau}=\pi_{\psi}(\mathbf{F}^{s}).
\end{equation}
The encoder maps visual observations into an intermediate representation $\mathbf{F}^{s}$, from which the policy head extracts information for trajectory planning.
The policy $\pi_{\psi}$ can follow regression-based~\citep{Kashyap2023Transfuser}, diffusion-based~\citep{liao2025diffusiondrive}, or scoring-based~\citep{feng2025rap3drasterizationaugmented} designs.
This formulation provides a common interface for studying visual representations across planners with different trajectory prediction mechanisms.

\paragraph{VFM-Guided Training.}
To incorporate powerful VFM representations into planner representation learning, we design a VFM-guided training pipeline.
During training, the same observation $\mathcal{O}$ is fed to a frozen VFM $T_{\phi}$ to obtain $\mathbf{F}^{t}=T_{\phi}(\mathcal{O})$.
These representations provide representation-level supervision for the planner encoder without being passed to the policy head $\pi_{\psi}$ or requiring additional annotations for planner training.
Only the original planner is kept at inference, leaving its deployed architecture and inference cost unchanged.
Unless otherwise specified, we use DVGT~\citep{zuo2025dvgt} as $T_{\phi}$, while the same pipeline also supports other VFMs~\citep{carion2025sam, wang2025vggt, simeoni2025dinov3, lin2025da3}.
Throughout the paper, DINOv3 refers to DINOv3-L unless otherwise specified.

\paragraph{Visual Representation Alignment.}
To support the spatial reasoning and scene understanding required for end-to-end autonomous driving, we design two objectives of ViRA to transfer high-quality VFM representations to the planner encoder: Spatial Dependency Alignment (SDA) and Scene Semantic Alignment (SSA).
SDA matches spatial attention distributions between $\mathbf{F}^{s}$ and $\mathbf{F}^{t}$, encouraging the planner to learn the spatial dependencies encoded by the VFM.
SSA aligns their global embeddings using a bidirectional contrastive objective, encouraging scene-level semantic consistency between the planner and the VFM.
The overall training objective is
\begin{equation}
\label{eq:total_loss}
\mathcal{L}
=
\mathcal{L}_{\mathrm{planner}}
+ \mathcal{L}_{\mathrm{align}},
\quad
\mathcal{L}_{\mathrm{align}}
=
\lambda_{\mathrm{SDA}}\mathcal{L}_{\mathrm{SDA}}
+
\lambda_{\mathrm{SSA}}\mathcal{L}_{\mathrm{SSA}}.
\end{equation}
where $\mathcal{L}_{\mathrm{planner}}$ is the original planner objective, including trajectory prediction and any auxiliary supervision used in the corresponding configuration, $\mathcal{L}_{\mathrm{align}}$ is the representation alignment objective combining SDA and SSA, and $\lambda_{\mathrm{SDA}}$ and $\lambda_{\mathrm{SSA}}$ are scalar weights.
Details of ViRA are provided in Appendix~\ref{app:method}.

\subsection{Benchmark and Evaluation}
\label{sec:exp_setup}

\textbf{NAVSIM}~\citep{dauner2024navsim} is a data-driven, non-reactive simulation and benchmarking framework for autonomous driving built on OpenScene~\citep{openscene2023} and nuPlan~\citep{karnchanachari2024towards}.
We use navtrain for training and evaluate models on two NAVSIM v2~\citep{cao2025navsimv2} benchmarks, navtest and navhard.
Both benchmarks adopt the rule-based planning metric EPDMS~\citep{li2025hydramdpplusplus}, whose sub-metrics include No-at-fault Collisions (NC), Drivable Area Compliance (DAC), Driving Direction Compliance (DDC), Traffic Light Compliance (TLC), Time-to-Collision (TTC), Ego Progress (EP), Lane Keeping (LK), History Comfort (HC), and Extended Comfort (EC).

\textbf{HUGSIM}~\citep{zhou2025hugsim} is a real-time, photorealistic closed-loop simulator that reconstructs 3D scenes from captured RGB images with 3D Gaussian Splatting and updates ego and actor states at each step from the planner output, so that surrounding vehicles react to the ego.
We evaluate planners trained only on NAVSIM in a zero-shot manner on scenarios grouped into four difficulty levels: Easy, Medium, Hard, and Extreme.
The evaluation metric is HD-Score, which weights route completion (RC) by per-step safety and comfort scores.
Beyond planning scores, we use mean Intersection-over-Union (mIoU) over seven BEV semantic classes to measure TransFuser's perception performance and assess the quality of its learned representations.
More details are provided in Appendix~\ref{app:impl}.
\begin{figure}[!htbp]
    \centering
    \includegraphics[width=0.9\linewidth]{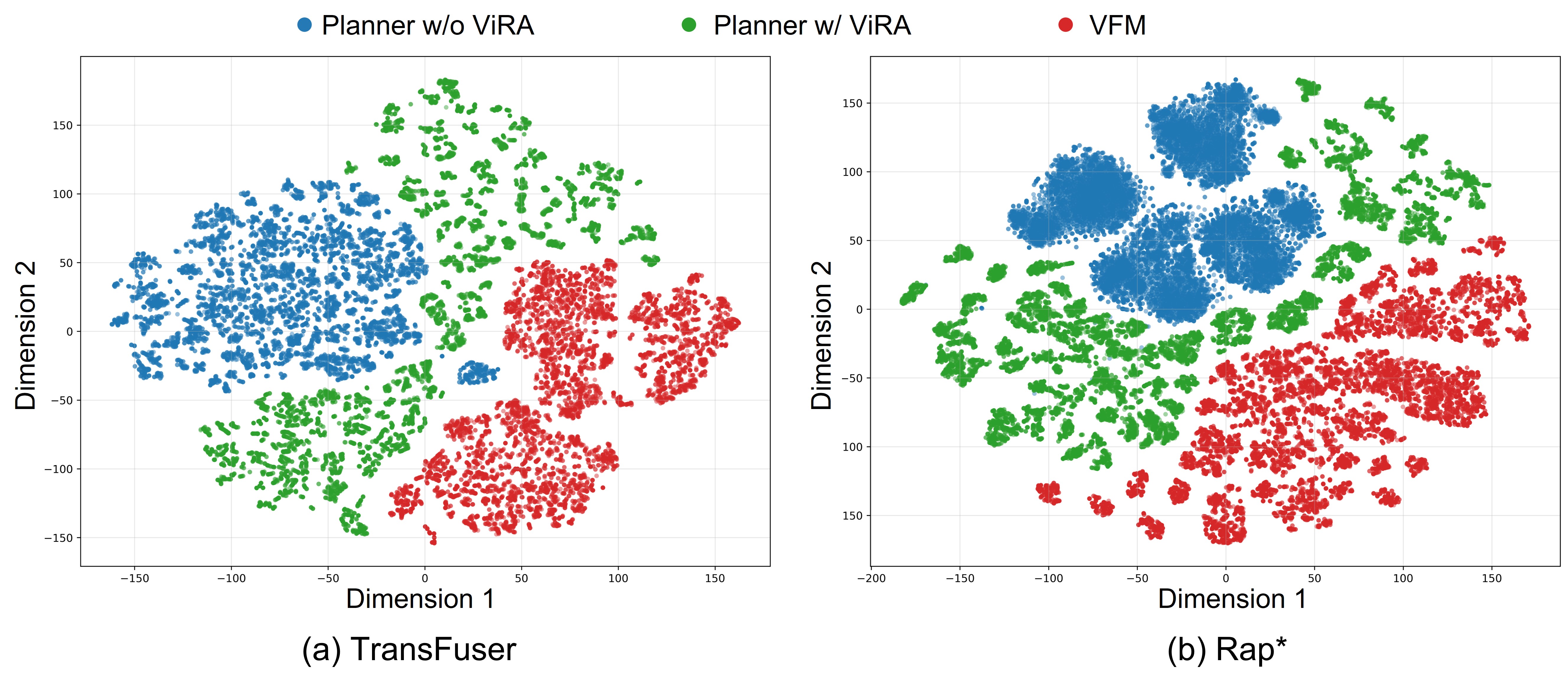}
    \caption{\textbf{t-SNE visualization of visual representations.} Planner representations without and with ViRA are shown alongside the reference VFM representations for (a) TransFuser and (b) Rap*. The visualization suggests that representation alignment with ViRA shifts planner representations toward those of the reference VFM in the t-SNE space.
}
    \label{fig:t-sne}
\end{figure}

\begin{table}[!ht]
	\caption{\textbf{Performance on the NAVSIM v2 navtest benchmark.}
	Rap$^\ast$ is our reimplementation with a different backbone from the official codebase and without the original data augmentation.
	For a fair comparison, all methods use camera input only, without LiDAR.}
	\label{tab:table1_vira_navtest}
	\setlength{\tabcolsep}{1.5pt}
	\renewcommand{\arraystretch}{1.05}
	\resizebox{\linewidth}{!}{
		\begin{tabular}{l|l|llllllllll}
			\toprule
			\textbf{Methods}
			& \textbf{Decoder}  
			& {\textbf{NC} $\uparrow$} 
			& {\textbf{DAC} $\uparrow$} 
			& {\textbf{DDC} $\uparrow$} 
			& {\textbf{TLC} $\uparrow$} 
			& {\textbf{EP} $\uparrow$} 
			& {\textbf{TTC} $\uparrow$} 
			& {\textbf{LK} $\uparrow$} 
			& {\textbf{HC} $\uparrow$} 
			& {\textbf{EC} $\uparrow$} 
			& \textbf{EPDMS} $\uparrow$ \\ 
			
			\midrule
			Human Agent 
			& \multicolumn{1}{c|}{-} 
			& {100} 
			& {100} 
			& {99.8} 
			& {100} 
			& {97.8} 
			& {100} 
			& {100}
			& {98.1} 
			& {90.1} 
			& {97.9}   
			\\
			
			\midrule
			\hline
			\multicolumn{12}{c}{\cellcolor[HTML]{E8E8E8}\textit{Perception-based}} 
			\\ 
			
			TransFuser~\citep{Kashyap2023Transfuser}
			& \multicolumn{1}{c|}{} 
			& {97.1} 
			& {89.6} 
			& {99.0} 
			& {99.8}
			& {97.9} 
			& {95.9}
			& {95.9}
			& {98.3}
			& {87.3} 
			& {83.6} \\ 
			
			TransFuser-ViRA 
			& \multicolumn{1}{c|}{\multirow{-2}{*}{Regression}}
			& {97.7} 
			& {94.1} 
			& {99.5} 
			& {99.8} 
			& {98.2} 
			& {96.7} 
			& {96.8} 
			& {98.3} 
			& {87.9} 
			& {\textbf{89.0} \textcolor[HTML]{C71585}{{\small +5.4}}} 
			
			\\ \midrule
			DiffusionDrive~\citep{liao2025diffusiondrive}
			& \multicolumn{1}{c|}{} 
			& {98.2} 
			& {95.9} 
			& {99.4} 
			& {99.8} 
			& {87.5} 
			& {97.3} 
			& {96.8} 
			& {98.3} 
			& {87.7}
			& {84.5} 
			\\ 
			DiffusionDrive-ViRA 
			& \multicolumn{1}{c|}{\multirow{-2}{*}{Diffusion}}
			& {98.2} 
			& {96.5} 
			& {99.6} 
			& {99.8} 
			& {98.2} 
			& {97.4} 
			& {97.4} 
			& {98.3} 
			& {88.0} 
			& {\textbf{91.9} \textcolor[HTML]{C71585}{{\small +7.4}}}
			
			\\ \midrule
			\hline
			\multicolumn{12}{c}{\cellcolor[HTML]{E8E8E8} \textit{Perception-free}} 

			\\
			Rap$^\ast$~\citep{feng2025rap3drasterizationaugmented}
			& \multicolumn{1}{c|}{} 
			& {91.6} 
			& {88.8} 
			& {93.5} 
			& {99.8} 
			& {87.2} 
			& {92.5} 
			& {90.9} 
			& {97.2} 
			& {56.7} 
			& {72.8}
			\\

			Rap$^\ast$-ViRA
			& \multicolumn{1}{c|}{\multirow{-2}{*}{Scoring}}
			& {97.1} 
			& {96.9} 
			& {99.2} 
			& {99.9} 
			& {87.5} 
			& {96.9} 
			& {96.2} 
			& {97.7} 
			& {48.1} 
			& {\textbf{83.7} \textcolor[HTML]{C71585}{{\small +10.9}}} 
			
			\\
			\bottomrule
	\end{tabular}}
\end{table}
\section{VFM-guided Visual Representations Do Improve Driving}
\label{sec:motivate}
\subsection{Driving Performance across Different Benchmarks}
\begin{table}[!h]
    \caption{\textbf{Performance on the NAVSIM v2 navhard benchmark.}
    Rap$^\ast$ is our reimplementation with a different backbone from the official codebase and without the original data augmentation.
    PDM-Closed plans with GT symbolic inputs, while all other methods use camera data only as input.}
    \label{tab:table2_vira_navhard}
    \setlength{\tabcolsep}{1.5pt}
	\renewcommand{\arraystretch}{1.05}
    \resizebox{\linewidth}{!}{
    \begin{tabular}{l|c|c|llllllllll}
    \toprule
    \textbf{Methods} 
    & {\textbf{Decoder}}
    & {\textbf{Stage}} 
    & {\textbf{NC} $\uparrow$} 
    & {\textbf{DAC} $\uparrow$}  
    & {\textbf{DDC} $\uparrow$}     
    & {\textbf{TLC} $\uparrow$}     
    & {\textbf{EP}  $\uparrow$}
    & {\textbf{TTC} $\uparrow$}  
    & {\textbf{LK} $\uparrow$}   
    & {\textbf{HC} $\uparrow$}    
    & {\textbf{EC} $\uparrow$}       
    & {\textbf{EPDMS} $\uparrow$}                      
    \\ 
    \midrule
    
    & {}
    & {S1}   
    & {94.4} 
    & {98.8} 
    & {100}  
    & {99.5}
    & {100}
    & {93.5} 
    & {99.3}
    & {87.7}
    & {36.0}
    &    
    
    \\
    \multirow{-2}{*}{PDM-Closed} 
    & \multicolumn{1}{c|}{\multirow{-2}{*}{-}}
    & {S2}
    & {90.5} 
    & {90.6}
    & {95.4} 
    & {98.4}
    & {100}
    & {98.4}
    & {74.2}
    & {91.9}
    & {29.7} 
    & {\multirow{-2}{*}{56.6}}      
    \\ 
    
    \midrule
    \hline
    \multicolumn{13}{c}{\cellcolor[HTML]{E8E8E8}\textit{Perception-based}}
    
    \\
    & {}
    & {S1}   
    & {96.4} 
    & {73.8} 
    & {95.6} 
    & {99.3} 
    & {94.8}
    & {95.6}
    & {88.0} 
    & {97.6}
    & {82.2}
    & {}                        
    
    \\
    \multirow{-2}{*}{TransFuser~\citep{Kashyap2023Transfuser}}
    & {}
    & {S2} 
    & {82.6} 
    & {62.8} 
    & {77.9} 
    & {98.2} 
    & {91.0}
    & {79.9}
    & {40.9}
    & {97.5}
    & {80.8} 
    & \multirow{-2}{*}{27.6} 
    
    \\
    \cmidrule{3-13}
    & {}
    & {S1} 
    & {97.4} 
    & {80.0} 
    & {98.7} 
    & {99.3}
    & {98.0}
    & {95.6}
    & {95.8}
    & {97.8}
    & {81.3} 
    & {}                   
    
    \\
    \multirow{-2}{*}{TransFuser-ViRA}
    & {\multirow{-4}{*}{Regression}}
    & {S2}   
    & {81.2} 
    & {73.9} 
    & {86.0} 
    & {98.0}
    & {97.0}
    & {78.3}
    & {47.3}
    & {96.1}
    & {71.1}
    & \multirow{-2}{*}{\textbf{31.7} \textcolor[HTML]{C71585}{{\small +4.1}}} 
    
    \\ \midrule
    & {}
    & {S1}   
    & {96.8} 
    & {86.0} 
    & {98.8} 
    & {99.3}
    & {84.0}
    & {95.8}
    & {96.7} 
    & {97.6}
    & {79.6} 
    &                         
    
    \\
    \multirow{-2}{*}{DiffusionDrive~\citep{liao2025diffusiondrive}}
    & {}
    & {S2}   
    & {80.1} 
    & {72.8} 
    & {84.4} 
    & {98.4}
    & {85.9}
    & {76.6}
    & {46.4} 
    & {96.3}
    & {72.8} 
    & \multirow{-2}{*}{30.5} 
    
    \\ \cmidrule{3-13}
    & {}
    & {S1} 
    & {96.4} 
    & {87.8} 
    & {99.2} 
    & {99.3}
    & {98.5}
    & {95.3} 
    & {96.4}
    & {97.6}
    & {78.7} 
    &                         
    
    \\
    \multirow{-2}{*}{DiffusionDrive-ViRA} 
    & {\multirow{-4}{*}{Diffusion}} 
    & {S2}   
    & {80.5} 
    & {74.1} 
    & {86.4} 
    & {98.4}
    & {97.5}
    & {77.2} 
    & {47.6}
    & {95.8}
    & {71.3} 
    & \multirow{-2}{*}{\textbf{35.8} \textcolor[HTML]{C71585}{{\small +5.3}}} 
    
    \\ \midrule
    \hline
    \multicolumn{13}{c}{\cellcolor[HTML]{E8E8E8}\textit{Perception-free}}
    \\ 
    & {}
    & {S1}
    & {89.4}
    & {71.8} 
    & {82.3} 
    & {99.6}
    & {86.6}
    & {89.6}   
    & {85.8}
    & {95.8}
    & {47.1}  
    &                            
    
    \\
    \multirow{-2}{*}{Rap$^\ast$~\citep{feng2025rap3drasterizationaugmented}}
    & {}
    & {S2}
    & {90.6}
    & {77.6}
    & {85.6}
    & {98.8}
    & {71.4}
    & {88.9}
    & {50.8}
    & {98.4}
    & {68.9}
    & \multirow{-2}{*}{32.8}
    
    \\
    \cmidrule{3-13}
    & {}
    & {S1}
    & {96.7}
    & {90.7}
    & {97.2}
    & {100.0}
    & {85.2}
    & {96.2}
    & {94.2}
    & {96.4}
    & {42.2}
    &                                      
                                 
    \\ 
    \multirow{-2}{*}{Rap$^\ast$-ViRA}
    & {\multirow{-4}{*}{Scoring}} 
    & {S2}   
    & {95.5}                
    & {87.6}                
    & {94.9}                
    & {99.6}
    & {69.3}
    & {94.5}                             
    & {55.3}
    & {98.7}
    & {68.3}                                     
    & \multirow{-2}{*}{\textbf{49.2} \textcolor[HTML]{C71585}{{\small +16.4}}}                  
    \\
    \bottomrule
    \end{tabular}}
\end{table}
\paragraph{Results on NAVSIM v2 navtest.} 
Table~\ref{tab:table1_vira_navtest} shows that VFM-guided representation alignment consistently improves driving performance across three representative planners.
With ViRA, TransFuser~\citep{Kashyap2023Transfuser}, DiffusionDrive~\citep{liao2025diffusiondrive}, and Rap$^\ast$~\citep{feng2025rap3drasterizationaugmented} achieve EPDMS gains of 5.4, 7.4, and 10.9 points over their respective baselines.
DAC and LK show substantial improvements, reflecting gains in drivable-area compliance and lane keeping, respectively. Both capabilities are closely tied to visual perception.
These improvements suggest that the benefits of VFM-guided visual representations extend across different planner architectures in standard driving scenarios.

\paragraph{Results on NAVSIM v2 navhard.}
The benefits extend to challenging driving scenarios: all three planners achieve higher EPDMS with ViRA on navhard (Table~\ref{tab:table2_vira_navhard}).
Notably, although Rap$^\ast$ already achieves the highest baseline EPDMS among the three planners, representation alignment further improves its score from 32.8 to 49.2, a gain of 16.4 points.
DAC and DDC also improve, reflecting better drivable-area compliance and driving direction compliance in challenging scenarios.
These improvements highlight the value of VFM-guided visual representations even for the strongest evaluated planner in challenging scenarios.
\begin{table}[t]
    \centering
    \caption{\textbf{Zero-shot Performance on the HUGSIM Benchmark.} Rap$^\ast$ is our reimplementation with a different backbone from the official codebase and without the original data augmentation.}
    \label{tab:hugsim_zero_shot}
    % \small
    \setlength{\tabcolsep}{2.0pt}
    \renewcommand{\arraystretch}{1.05}
    \resizebox{\linewidth}{!}{%
    \begin{tabular}{l|cc cc cc cc|cl}
        \toprule
        \multirow{2}{*}{Method}
        & \multicolumn{2}{c}{Easy}
        & \multicolumn{2}{c}{Medium}
        & \multicolumn{2}{c}{Hard}
        & \multicolumn{2}{c|}{Extreme}
        & \multicolumn{2}{c}{Overall} \\
        \cmidrule(lr){2-3}
        \cmidrule(lr){4-5}
        \cmidrule(lr){6-7}
        \cmidrule(lr){8-9}
        \cmidrule(lr){10-11}
        & RC & HD-Score
        & RC & HD-Score
        & RC & HD-Score
        & RC & HD-Score
        & RC & HD-Score \\
        
        \midrule
        \hline
        \multicolumn{11}{c}{\cellcolor[HTML]{E8E8E8}\textit{Perception-based}} \\

        TransFuser~\citep{Kashyap2023Transfuser}
        & 66.8 & 56.7
        & 28.1 & 11.6
        & 24.5 & 9.2
        & 26.2 & 10.9
        & 36.4 & 22.1 \\

        TransFuser-ViRA
        & 72.7 & 62.6
        & 31.9 & 15.3
        & 27.8 & 11.4
        & 25.5 & 9.9
        & 39.5  & \textbf{24.8} \textcolor[HTML]{C71585}{{\small +2.7}} \\

        \midrule

        DiffusionDrive~\citep{liao2025diffusiondrive}
        & 68.9 & 56.8
        & 29.8 & 12.4
        & 25.9 & 9.7
        & 25.7 & 9.8
        & 37.6 & 22.2
        \\

        DiffusionDrive-ViRA
        & 77.5 & 69.3
        & 37.7 & 20.7
        & 30.0 & 12.0
        & 27.9 & 12.3
        & 43.3 & \textbf{28.6} \textcolor[HTML]{C71585}{{\small +6.4}}
        \\

        \midrule
        \hline
        \multicolumn{11}{c}{\cellcolor[HTML]{E8E8E8}\textit{Perception-free}}  \\
        Rap$^\ast$~\citep{feng2025rap3drasterizationaugmented}
        & 11.8 & 10.6
        & 11.1 & 9.9 
        & 7.8 & 6.3
        & 4.1 & 2.3
        & 8.7 & 7.3
        \\

        Rap$^\ast$-ViRA
        & 18.1 & 15.9
        & 16.7 & 14.6
        & 10.5 & 8.2
        & 4.3 & 2.3
        & 12.4 & \textbf{10.3} \textcolor[HTML]{C71585}{{\small +3.0}}
        \\
        \bottomrule
    \end{tabular}%
    }
\end{table}
\begin{figure}[!htbp]
    \centering
    \includegraphics[width=1.0\linewidth]{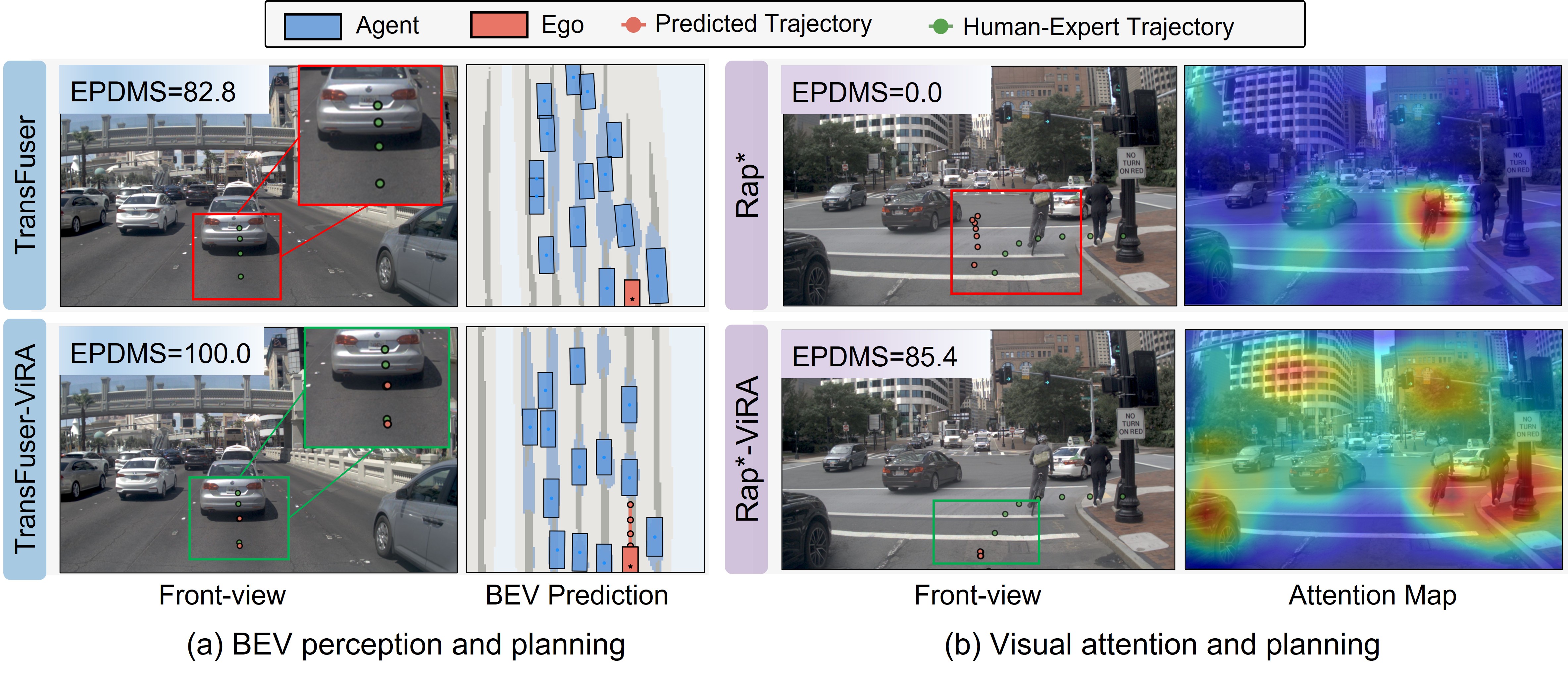}
    \caption{\textbf{Qualitative results showing improved perception and planning with ViRA.}}
    \label{fig:quali}
\end{figure}
\paragraph{Zero-shot Results on HUGSIM.}
To assess whether the gains extend beyond NAVSIM, we evaluate NAVSIM-trained planners on the closed-loop HUGSIM benchmark~\citep{zhou2025hugsim} without further training.
As shown in Table~\ref{tab:hugsim_zero_shot}, ViRA improves the overall HD-Score of TransFuser, DiffusionDrive and Rap$^\ast$ by 2.7, 6.4 and 3.0 points, respectively.
Together with the NAVSIM results, these zero-shot gains support our finding that VFM-guided visual representations do improve driving performance across different benchmarks and evaluation settings.
\subsection{Representation Quality Improves Measurably}
To examine whether the driving gains are accompanied by improved perception, we analyze BEV segmentation and attention maps.
Figure~\ref{fig:delta_results_miou_epmds}(a) shows that ViRA improves TransFuser's BEV segmentation mIoU by 0.5--3.0 points across the evaluated VFM targets.
The t-SNE visualizations in Figure~\ref{fig:t-sne} suggest that ViRA shifts the representations of TransFuser and Rap$^\ast$ toward those of the DVGT target in the t-SNE embedding space.  
The qualitative examples in Figure~\ref{fig:quali} further illustrate improvements in BEV predictions and planned trajectories, together with changes in attention to surrounding road users.
These results provide evidence that alignment with high-quality VFM representations improves perception alongside driving performance.

\begin{table}[!h]
\centering
\caption{\textbf{VFMs dependency on BEV segmentation of TransFuser.} Static objects and pedestrians are omitted from the class-wise results; mIoU is computed over all seven classes.}
\label{tab:bev_miou_random}
\small
\resizebox{\linewidth}{!}{
\begin{tabular}{l|clcccl}
\toprule
\textbf{VFM}
& \textbf{Background $\uparrow$}
& \textbf{Road $\uparrow$} 
& \textbf{Walkway $\uparrow$}
& \textbf{Centerline $\uparrow$}
& \textbf{Vehicle $\uparrow$}
& \textbf{mIoU $\uparrow$} \\

\midrule
Baseline
& 85.4
& 62.3
& 56.3
& 23.6
& 27.1
& 36.4 \\

\midrule

Random-init
& 82.4
& 58.3
& 47.2
& 12.5
& 14.7 
& 30.7 \textcolor[HTML]{0073CF}{{\footnotesize -5.7}}

\\

DVGT~\citep{zuo2025dvgt}
& 86.9
& 64.6
& 59.3
& 28.9
& 33.4
& \textbf{39.1} \textcolor[HTML]{C71585}{{\footnotesize +2.7}}\\

\bottomrule
\end{tabular}}
\end{table}
\begin{table}[!ht]
\caption{\textbf{Effect of different VFM teachers on TransFuser and Rap*.} Performance on the NAVSIM v2 navtest benchmark.}
\label{tab:navtest_random_teacher}
\small
\resizebox{\linewidth}{!}{
\begin{tabular}{l|l|llllll}
\toprule 
{\textbf{Methods}}
& {\textbf{VFMs}}
& {\textbf{NC} $\uparrow$} 
& {\textbf{DAC} $\uparrow$} 
& {\textbf{DDC} $\uparrow$} 
& {\textbf{TLC} $\uparrow$} 
& {\textbf{LK} $\uparrow$} 
& \textbf{EPDMS} $\uparrow$ 
\\ 
\midrule
\hline
\multicolumn{8}{c}{\cellcolor[HTML]{E8E8E8}\textit{Perception-based}} 
\\ 
{}
& {Baseline}
& {97.1} 
& {89.6} 
& {99.0} 
& {99.8} 
& {95.9} 
& {83.6} 
\\ 

{}
& {Random-init}
& {96.8}
& {90.2}
& {99.0}
& {99.7}
& {95.5}
& 82.0 \textcolor[HTML]{0073CF}{{\footnotesize -1.6}}

\\
\multirow{-3}{*}{TransFuser~\citep{Kashyap2023Transfuser}}
& {DVGT~\citep{zuo2025dvgt}}
& {97.7} 
& {94.1} 
& {99.5} 
& {99.8} 
& {96.8} 
& {\textbf{89.0} \textcolor[HTML]{C71585}{{\footnotesize +5.4}}} 
\\
\midrule
\hline
\multicolumn{8}{c}{\cellcolor[HTML]{E8E8E8}\textit{Perception-free}} 
\\ 
{}
& {Baseline}
& {91.6}
& {88.8}
& {93.5}
& {99.8}
& {90.9}
& {72.8}  \\

{}
& {Random-init}
& {94.6}
& {88.1}
& {95.1}
& {99.8}
& {93.1}
& {71.1} \textcolor[HTML]{0073CF}{{\footnotesize -1.7}}\\

\multirow{-3}{*}{Rap*~\citep{feng2025rap3drasterizationaugmented}}
& {DVGT~\citep{zuo2025dvgt}}
& {97.1}
& {96.9}
& {99.2}
& {99.9}
& {96.2}
& {\textbf{83.7} \textcolor[HTML]{C71585}{{\footnotesize +10.9}}}
\\

\bottomrule
\end{tabular}}
\end{table}
\subsection{The Gains Require a Pre-trained Target}
To test whether the gains depend on VFM pre-training, we replace the pre-trained target with a randomly initialized one while retaining the same alignment objectives.
As shown in Table~\ref{tab:bev_miou_random}, random-target alignment reduces TransFuser's BEV segmentation mIoU from 36.4 to 30.7, whereas alignment with pre-trained DVGT increases it to 39.1.
The same pattern holds for driving performance: random-target alignment lowers EPDMS by 1.6 points for TransFuser and 1.7 points for Rap$^\ast$, while the pre-trained target improves both planners (Table~\ref{tab:navtest_random_teacher}).
These results support the importance of pre-trained visual knowledge: applying the same alignment objectives to random representations does not reproduce the gains observed with pre-trained VFM representations.

\subsection{Gains without Additional Inference Cost}
\begin{wrapfigure}{r}{0.48\linewidth}
\centering
\vspace{-5pt}
\includegraphics[width=\linewidth]{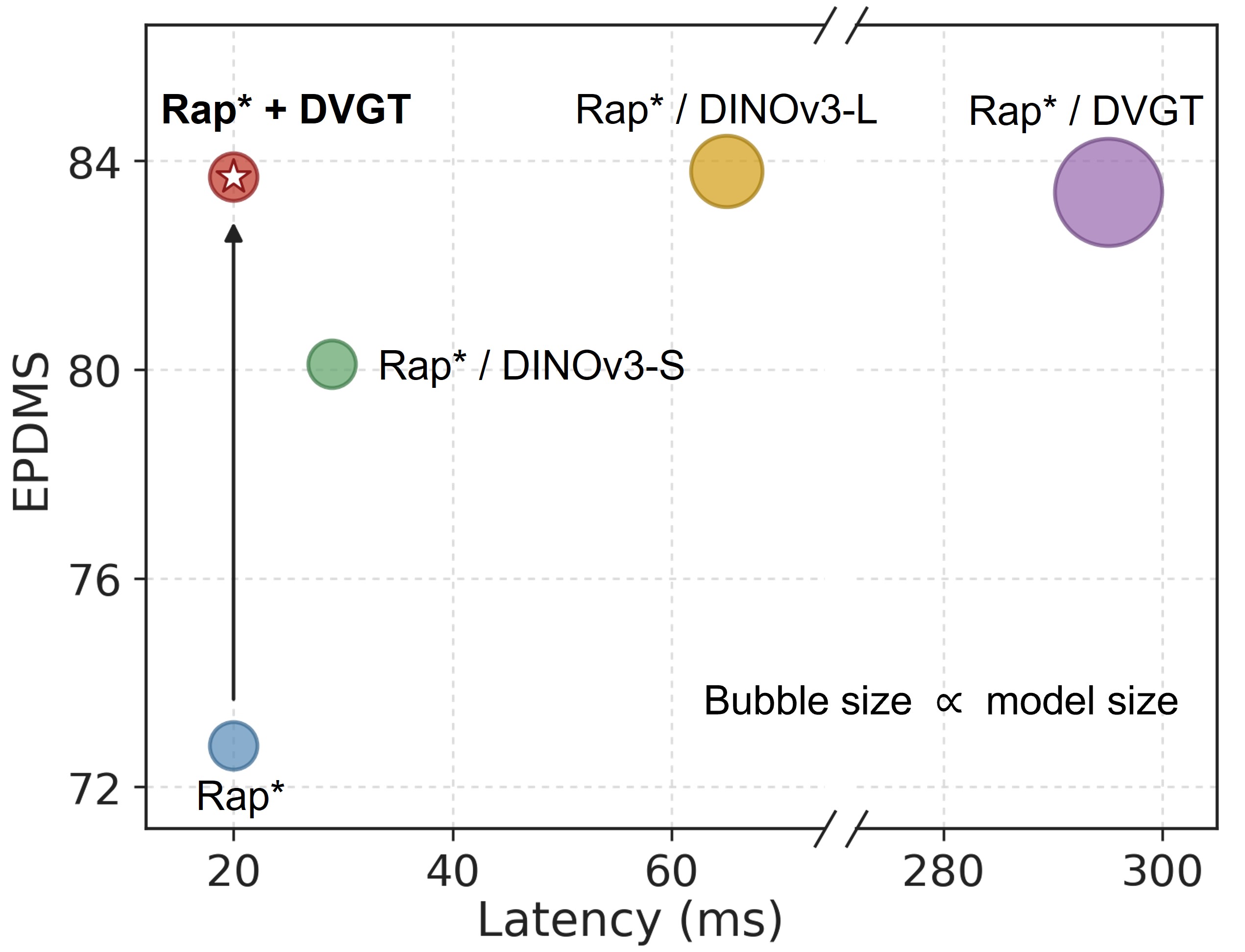}
\captionsetup{skip=3pt,font=small}
\caption{\textbf{ViRA (star) vs.\ backbone replacement (circles).}
Bubble size indicates parameter count.
}

\label{fig:inference_performance}
\vspace{-5pt}
\end{wrapfigure}
A practical question is whether these representation gains require replacing the planner encoder with a larger VFM at inference.
Figure~\ref{fig:inference_performance} compares ViRA with direct VFM backbone replacement in terms of planning performance, inference latency, and parameter count. 
Latency is measured on a single NVIDIA H100 GPU.
DVGT-guided alignment improves EPDMS from 72.8 to 83.7 while retaining the baseline latency of 20~ms and parameter count of 69M. 
This nearly matches the strongest evaluated backbone replacement, DINOv3-L, which achieves 83.8 EPDMS but requires 65~ms and 351M parameters.
Since the VFM and alignment heads are used only during training, ViRA enables the planner to benefit from pre-trained visual representations without changing its deployed architecture or increasing inference cost.

\section{Planning Gains Depend on Both VFMs and Planners}
\label{sec:analysis}

\subsection{The choice of VFM target matters}
\begin{table}[!ht]\caption{\textbf{Rap$^\ast$~\citep{feng2025rap3drasterizationaugmented} with DINOv3~\citep{simeoni2025dinov3} backbones.} Performance on the NAVSIM v2 navtest benchmark under different VFM teachers.}
\label{tab:dinov3_backbone}
\resizebox{\linewidth}{!}{
\begin{tabular}{l|c|lllllllllll}
\toprule
\textbf{Backbone}  
&{\textbf{Teachers}}
& {\textbf{NC} $\uparrow$} 
& {\textbf{DAC} $\uparrow$} 
& {\textbf{DDC} $\uparrow$} 
& {\textbf{TLC} $\uparrow$} 
& {\textbf{LK} $\uparrow$} 
& \textbf{EPDMS} $\uparrow$ \\ 

\midrule
DINOv3-vit16l
& \multicolumn{1}{c|}{-}
& 97.2
& 96.4
& 98.9
& 99.9
& 96.1
& 83.8 
\\

DINOv3-vit16l
& VGGT~\citep{wang2025vggt}
& 98.5
& 97.3
& 99.3
& 100.0
& 96.7
& \textbf{86.3} \textcolor[HTML]{C71585}{{\small +2.5}}

\\
\bottomrule
\end{tabular}}

\end{table}
As shown in Figure~\ref{fig:teaser_overview}(c-ii), changing only the representation alignment target for the perception-free Rap$^\ast$ yields EPDMS scores ranging from 79.4 to 84.9, a spread of 5.5 points.
This difference is comparable to the gains from many method-level improvements.
These results highlight the importance of alignment target selection, with DINOv3, which learns from more general data, providing the largest planning gain in this setting.
To further examine whether VFMs with different pre-training objectives provide complementary representations, we use Rap$^\ast$ with a DINOv3-L backbone as the baseline and introduce alignment to VGGT, a VFM trained for 3D geometry.

As shown in Table~\ref{tab:dinov3_backbone}, VGGT alignment improves EPDMS by 2.5 points, from 83.8 to 86.3.
Thus, even when the backbone already provides strong semantic representations, alignment to a geometry-focused VFM can further improve planning performance.
These results suggest that alignment target selection deserves careful consideration, as its benefits may depend on the complementarity between the target VFM and the planner's existing representations.
\subsection{Auxiliary Supervision Shapes Sensitivity to VFM Targets}
\begin{wrapfigure}{r}{0.48\linewidth}
    \centering
    \vspace{-10pt}
    \includegraphics[width=\linewidth]{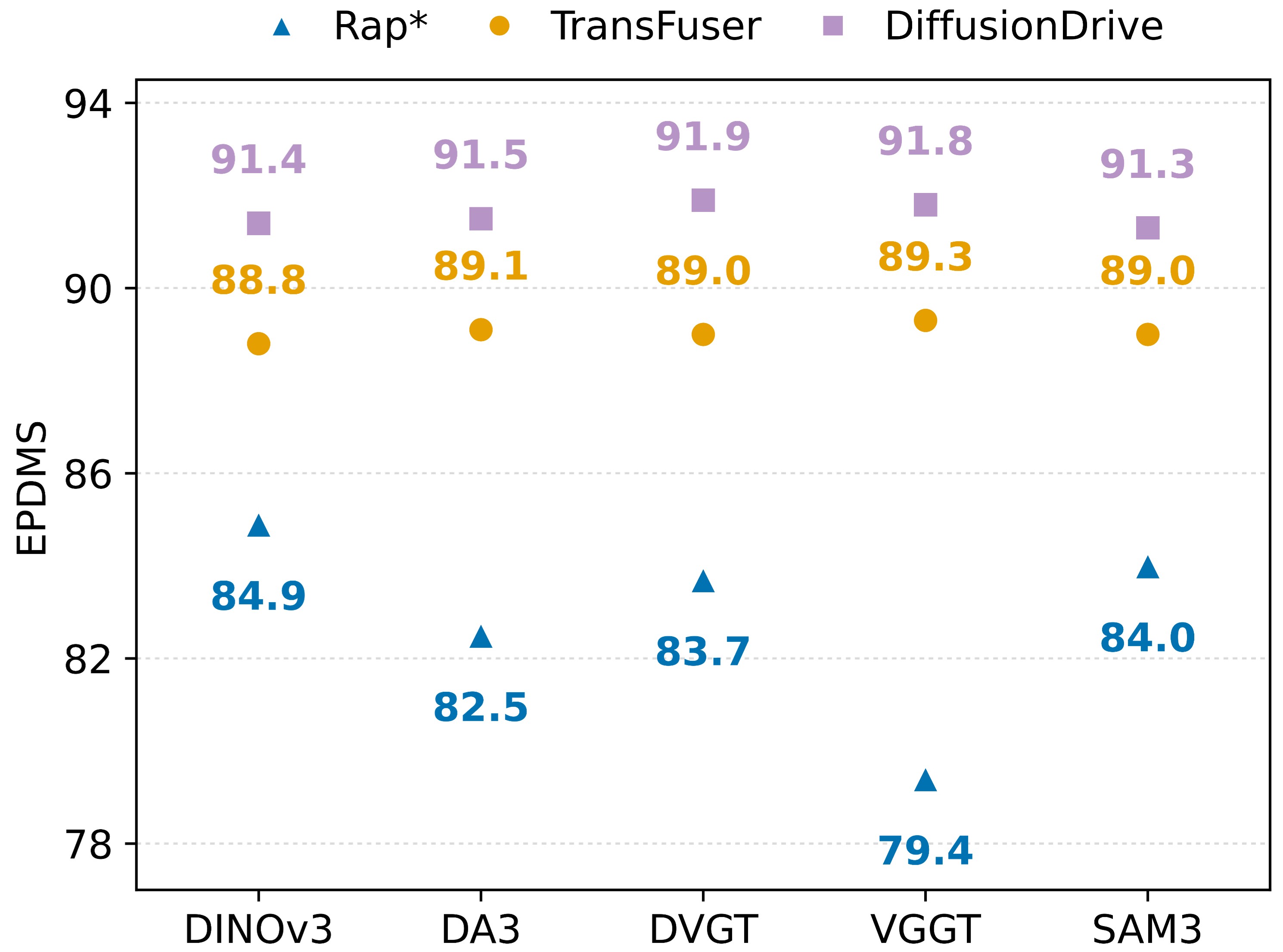}
    \captionsetup{skip=3pt,font=small}
    \caption{\textbf{Different VFM targets across planners.}
    }
    
    \label{fig:different_vfms_across_planners}
    \vspace{-10pt}
\end{wrapfigure}
Extending the experiments from the preceding subsection to perception-based planners, we find that TransFuser and DiffusionDrive are less sensitive to the choice of VFM target.
Across VFM targets, EPDMS varies by only 0.5 points for TransFuser (88.8 to 89.3) and 0.6 points for DiffusionDrive (91.3 to 91.9), compared with 5.5 points for the perception-free Rap$^\ast$.
One important difference between these planners is the use of auxiliary perception supervision.
TransFuser and DiffusionDrive learn planner representations with BEV segmentation and agent detection supervision, whereas Rap$^\ast$ does not.
We therefore hypothesize that auxiliary perception supervision reduces the sensitivity of the planner to VFM target selection.

\begin{figure}[!htbp]
    \centering
    \includegraphics[width=1\textwidth]{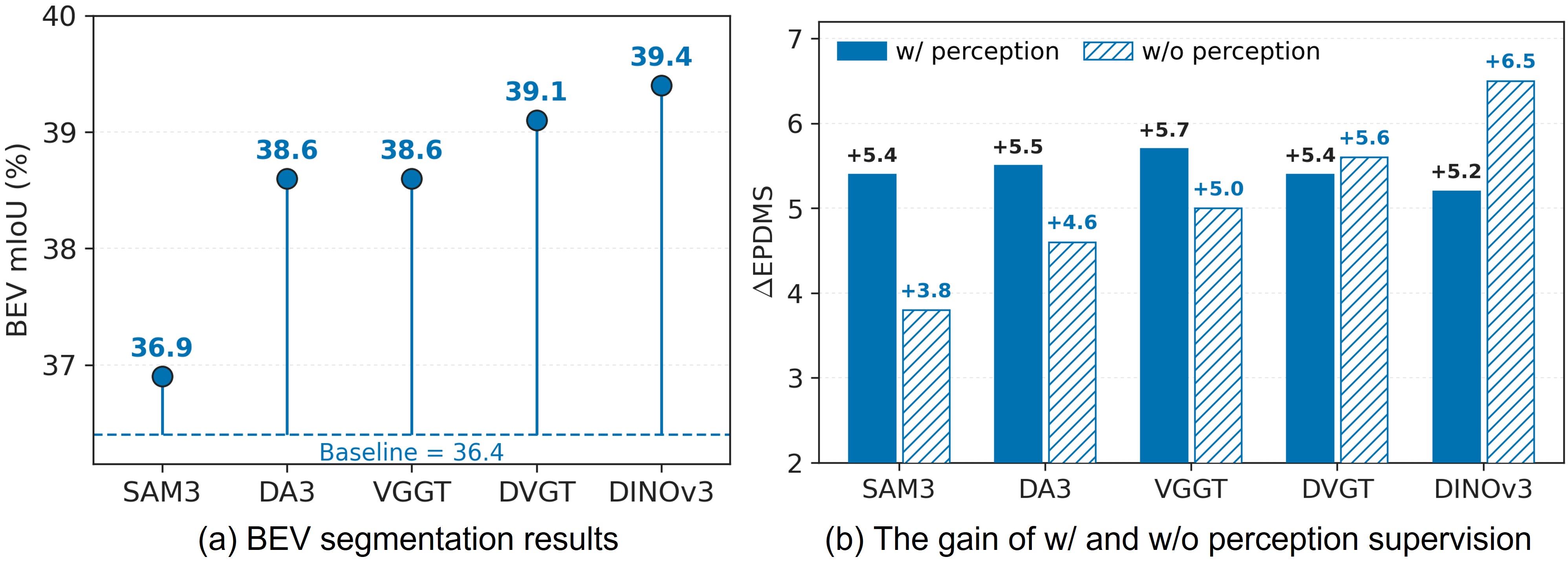}
    \caption{\textbf{Auxiliary perception supervision narrows the spread of planning gains across VFM targets.}
    (a) BEV mIoU across VFM targets; the dashed line marks the baseline (36.4).
    (b) Planning gain $\Delta$EPDMS with and without auxiliary perception supervision
    (baselines 83.6 and 82.6, respectively).}
    \label{fig:delta_results_miou_epmds}
\end{figure}

To test this hypothesis, we compare TransFuser~\citep{Kashyap2023Transfuser} with and without auxiliary perception supervision across five alignment targets: DVGT~\citep{zuo2025dvgt}, VGGT~\citep{wang2025vggt}, DA3~\citep{lin2025da3}, DINOv3~\citep{simeoni2025dinov3}, and SAM3~\citep{carion2025sam}.
As shown in Figure~\ref{fig:delta_results_miou_epmds}(b), EPDMS ranges from 88.8 to 89.3 with auxiliary supervision.
Removing this supervision expands the range to 86.4 to 89.1, increasing the spread from 0.5 to 2.7 points.
Examining the endpoints of these ranges provides further insight: after removing auxiliary supervision, the maximum score remains nearly unchanged (89.3 to 89.1), whereas the minimum drops substantially (88.8 to 86.4).
This indicates that auxiliary supervision mainly compensates for the less effective targets.
Across these targets, TransFuser's BEV segmentation results under auxiliary supervision show a broadly similar ordering to its planning results without auxiliary supervision.
Overall, these controlled comparisons show that auxiliary perception supervision reduces TransFuser's sensitivity to VFM target selection. 
\begin{table}[!ht]
    \caption{\textbf{Comparison with state-of-the-art methods on the NAVSIM v2 navtest benchmark.} ViRA-Diffusion is DiffusionDrive~\citep{liao2025diffusiondrive} trained with ViRA and without auxiliary perception supervision.}
    \label{tab:navtest_com}
    \centering
    \footnotesize
    \setlength{\tabcolsep}{1.3pt}
    \begin{tabular}{l|cccccccccc}
    \toprule
    \textbf{Method}  
    & {\textbf{NC} $\uparrow$} 
    & {\textbf{DAC} $\uparrow$} 
    & {\textbf{DDC} $\uparrow$} 
    & {\textbf{TLC} $\uparrow$} 
    & {\textbf{EP} $\uparrow$} 
    & {\textbf{TTC} $\uparrow$} 
    & {\textbf{LK} $\uparrow$} 
    & {\textbf{HC} $\uparrow$} 
    & {\textbf{EC} $\uparrow$} 
    & \textbf{EPDMS} $\uparrow$ \\ 
    
    \midrule
    Epona~\citep{zhang2025epona}
    & 97.1
    & 95.7
    & 99.3
    & 99.7
    & 88.6
    & 96.3
    & 97.0
    & 98.0
    & 67.8
    & 85.1 \\
    
    DiffusionDriveV2~\citep{zou2025diffusiondrivev2}
    & 97.7
    & 96.6
    & 99.2
    & 99.8
    & 88.9
    & 97.2
    & 96.0
    & 97.8
    & 91.0
    & 87.5 
    \\
      
    Latent-WAM~\citep{wang2026latent-wam}
    & 98.1
    & 97.3
    & 99.6
    & 99.8
    & 87.7
    & 97.3
    & 97.6
    & 98.1
    & 87.3
    & 89.3
    \\
    
    DriveFuture~\citep{hong2026drivefuture}
    & 98.8
    & 99.1
    & 99.6
    & 99.9
    & 86.6
    & 98.4
    & 96.4
    & 98.3
    & 74.8
    & 89.9
    \\
    
    SparseDriveV2~\citep{sun2026sparsedrivev2}
    & 98.1
    & 98.1
    & 99.6
    & 99.8
    & 91.1
    & 97.3
    & 96.9
    & 98.2
    & 78.4
    & 90.1
    \\
    
    Discrete-WAM~\citep{yao2026discrete}
    & 98.5
    & 98.2
    & 99.7
    & 99.8
    & 90.5
    & 97.9
    & 97.2
    & 98.3
    & 78.1
    & 90.4
    \\

    \rowcolor[HTML]{EDE4F1}
    \textbf{ViRA-Diffusion}
    & 98.3
    & 96.8
    & 99.5
    & 99.8
    & 98.2
    & 97.5
    & 97.4
    & 98.3
    & 88.0
    & \textbf{92.3}
    \\
    \bottomrule
    \end{tabular}
\end{table}

\subsection{Practical Implications}
The preceding analysis suggests that a planner can maintain comparable planning performance without auxiliary perception supervision when aligned with a suitable VFM target.
Building on these findings, we develop ViRA-Diffusion, a diffusion-based planner~\citep{liao2025diffusiondrive} trained with ViRA using DINOv3 as the target and no auxiliary perception supervision.
As shown in Table~\ref{tab:navtest_com}, the resulting ViRA-Diffusion achieves 92.3 EPDMS on NAVSIM v2 navtest, comparing favorably with the recent results included in our comparison.
This result illustrates the practical value of our analysis in guiding the joint choice of VFM targets and supervision strategies for end-to-end planners.

\section{Related Work}
\label{sec:related}

\paragraph{End-to-End Autonomous Driving.}
End-to-end autonomous driving learns planning and control directly from sensory inputs within a unified framework~\citep{hu2023Uniad, yu2025distilldrive, sun2025sparsedrive}.
Joint optimization can mitigate cascading errors between separately trained modules and facilitate learning from large-scale driving data, making end-to-end autonomous driving a promising route toward scalable real-world deployment~\citep{chen2024end-to-end,jia2024bench2drive, guan2026planning}.
Recent studies focus on improving policy heads~\citep{liao2025diffusiondrive, 2025rad, zheng2026unleashing, yang2026drivemoe}, introducing auxiliary perception supervision~\citep{strong2026learning, chen2024vadv2, xing2025goalflow}, and scaling training data through synthetic scene generation and augmentation~\citep{yang2023llm4drive,zhou2025nexus, xu2025challenger, li2025mtgs, tian2025simscale, li2026optimization,li2026world}.
Despite this progress, how visual representation quality contributes to planning gains remains insufficiently understood.
Some recent methods enhance visual representations for driving with semantic and geometric priors from pre-trained VFMs~\citep{zheng2025world4drive, wang2026latent-wam}.
These methods demonstrate benefits within specific systems, but provide limited guidance on how the gains vary across VFM targets and planner designs.
Complementing these directions, we study when better visual representations improve planning performance.

\paragraph{Representation Learning for Driving.}
Learning visual representations that capture scene semantics and spatial structure is crucial for end-to-end autonomous driving~\citep{chen2024end-to-end}.
To enrich these representations, recent driving studies turn to VFMs pre-trained on large-scale visual data, drawing on their capabilities in semantics~\citep{kirillov2023segment, carion2025sam}, depth~\citep{yang2024da1, yang2024da2, lin2025da3}, and 3D geometry~\citep{zuo2025dvgt, wang2024dust3r, wang2025vggt}.
These VFMs are used through visual backbone replacement~\citep{feng2025rap3drasterizationaugmented, naeinian2026zero}, auxiliary supervision from VFM predictions~\citep{jia2025drivevggt, strong2026learning}, or feature integration within task-specific architectures and training objectives~\citep{zheng2025world4drive, wang2026latent-wam, Wang2026vggdrive, xie2026latentvla}.
Recent studies show that representation alignment, which supervises intermediate features with frozen visual representations, can improve training efficiency and task performance~\citep{yu2024repa, li2025sf}.
iREPA~\citep{singh2025irepa} further shows that the spatial structure of target representations predicts image generation quality better than their ImageNet classification accuracy.
In end-to-end autonomous driving, how planner architecture and auxiliary perception supervision shape the benefits of VFM-guided representations remains unclear.
We use representation alignment as a controlled experimental interface to study these interactions across planners and VFM targets.

\section{Conclusion and Outlook}
\label{sec:conclusion}
In this paper, we systematically study when VFM-guided visual representations improve end-to-end autonomous driving, using ViRA as an experimental interface that keeps the planner's architecture and inference cost unchanged.
Experiments across planners and benchmarks show consistent driving gains, with benefits extending to zero-shot closed-loop evaluation.
VFM target selection affects these gains, and alignment to a different VFM can further improve planners that already use pre-trained VFM backbones.
Controlled comparisons on TransFuser further show that auxiliary perception supervision narrows performance differences across VFM targets.
These results highlight the importance of considering VFM target selection and planner supervision jointly.
Guided by this analysis, we develop ViRA-Diffusion with DINOv3 alignment and no auxiliary perception supervision, achieving 92.3 EPDMS on NAVSIM v2 navtest.
Future work will examine whether these findings extend to additional planner architectures and other physical intelligence tasks involving decision-making.
\subsubsection*{Acknowledgments}

We would like to thank Fanghua Yu, Jiazhi Yang, and other members of OpenDriveLab for their valuable discussions and technical support.

\bibliography{reference}
\bibliographystyle{bib_style}

\clearpage  
\appendix
\vskip8pt
\noindent\textbf{\textit{\LARGE Appendix}}
\vskip8pt

\setcounter{figure}{0}
\setcounter{table}{0}
\setcounter{equation}{0}

\renewcommand{\thefigure}{S.\arabic{figure}}
\renewcommand{\thetable}{S.\arabic{table}}
\renewcommand{\theequation}{S.\arabic{equation}}

\renewcommand{\theHfigure}{S.\arabic{figure}}
\renewcommand{\theHtable}{S.\arabic{table}}
\renewcommand{\theHequation}{S.\arabic{equation}}

The appendix is organized as follows. 
Appendix~\ref{app:method} provides the method details.
Appendix~\ref{app:impl} provides evaluation and implementation details.
Appendix~\ref{app:add_exp} reports additional quantitative results.
Appendix~\ref{app:qual} presents additional qualitative visualizations and failure cases.
Appendix~\ref{app:limitation} discusses limitations and broader impact.
Appendix~\ref{app:license} lists the licenses of the assets used in this work.

\section{VFM-guided Representation Alignment}
\label{app:method}
We present the ViRA framework as follows.
Section~\ref{sec:preli} defines the notation and problem setup.
Appendix~\ref{sec:pipeline} provides an overview of the ViRA framework for E2E planners.
Appendix~\ref{sec:LA} introduces Spatial Dependency Alignment (SDA) and Scene Semantic Alignment (SSA).
Appendix~\ref{sec:objective} summarizes the training objective.

\begin{figure}[htbp]
  \centering
  \includegraphics[width=1.0\textwidth]{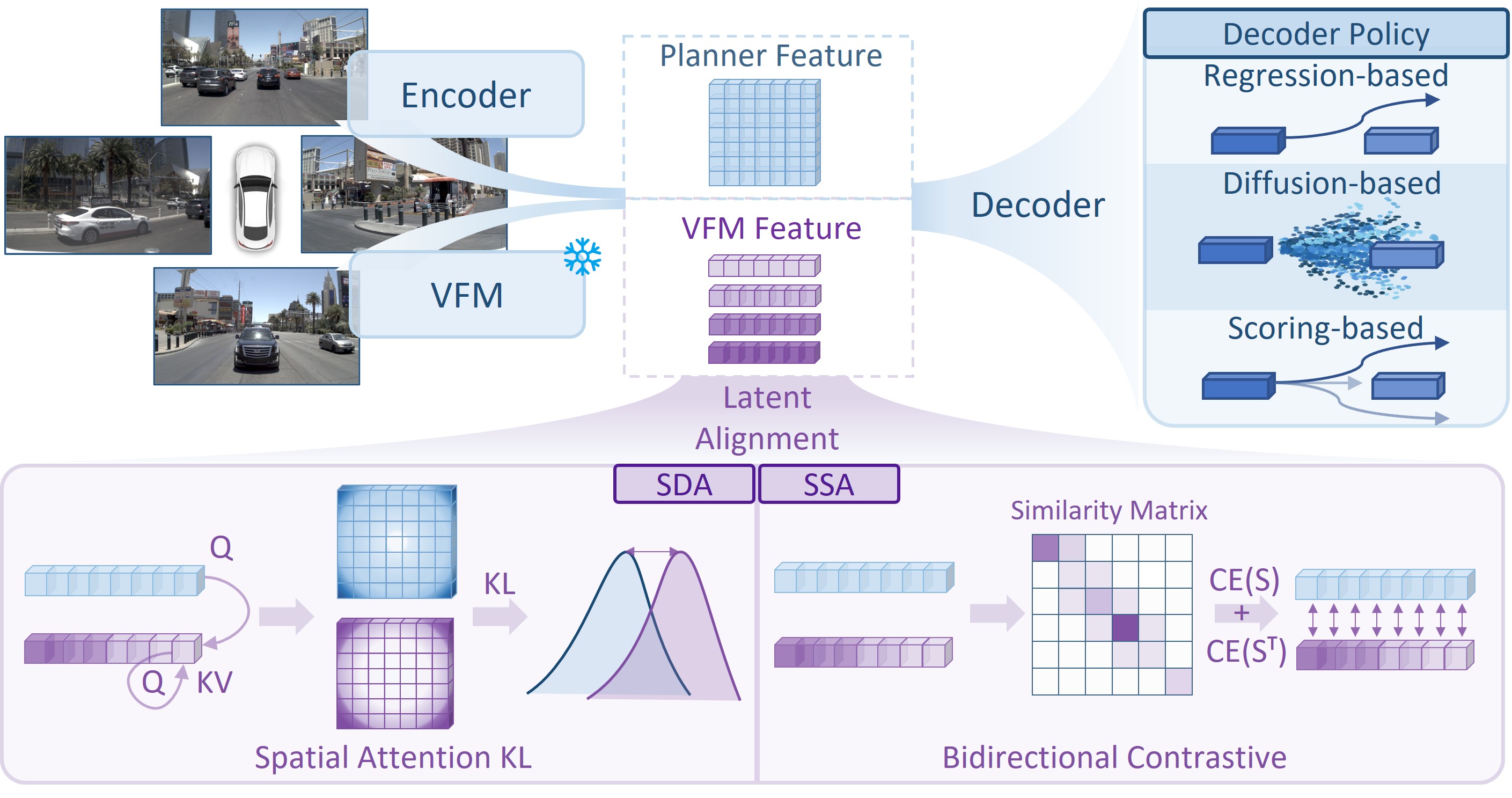}
  \caption{\textbf{Pipeline of ViRA.}
  Multi-view images are encoded by both the planner encoder and a frozen VFM target, producing student and VFM representations.
  The VFM representations supervise latent alignment only, without being injected into the policy head.
  The policy head operates solely on planner representations, and the VFM branch is removed after training, yielding no inference overhead.
  Q, K, and V denote query, key, and value; KL denotes Kullback--Leibler divergence; CE denotes cross-entropy.
  }
  \label{fig:pipeline}
  \vspace{-10pt}
\end{figure}

\subsection{VFM-Guided E2E Planner}
\label{sec:pipeline}

 As illustrated in Figure~\ref{fig:pipeline}, given a single-frame multi-view observation $\mathcal{O}$, a conventional E2E planner first extracts an intermediate representation $\mathbf{F}^{s}$ using the planner encoder $E_{\theta}$, and then predicts the future trajectory $\hat{\tau}$ with a driving policy $\pi_{\psi}$:
\begin{equation}
\label{eq:planner}
\mathbf{F}^{s}=E_{\theta}(\mathcal{O}), 
\qquad 
\hat{\tau}=\pi_{\psi}(\mathbf{F}^{s}).
\end{equation}
Here, $\pi_{\psi}$ denotes a general policy head, which can be instantiated by different E2E planner designs, such as regression-based~\citep{Kashyap2023Transfuser}, diffusion-based~\citep{liao2025diffusiondrive}, or scoring-based~\citep{feng2025rap3drasterizationaugmented} policies.

To improve the representation quality of the planner encoder, we additionally feed the same observation $\mathcal{O}$ into a pre-trained VFM $T_{\phi}$ to obtain a VFM representation $\mathbf{F}^{t}=T_{\phi}(\mathcal{O})$, where $\phi$ is frozen throughout training.
The VFM representation is used only to compute the representation alignment loss and is then discarded, without being passed to the policy head.
Unless otherwise specified, we use DVGT~\citep{zuo2025dvgt} as the default VFM target $T_{\phi}$, a VFM pre-trained on autonomous driving data, while our framework is also compatible with other VFMs~\citep{carion2025sam, wang2025vggt, simeoni2025dinov3, lin2025da3}.
Importantly, the VFM and alignment heads are used only during training and discarded during inference, keeping the original planner architecture and inference pipeline unchanged with no additional inference cost.
Since ViRA uses frozen VFM representations as supervision, it requires no extra annotations.

\subsection{Latent Representation Alignment}
\label{sec:LA}
To transfer driving-relevant knowledge from VFMs to planner encoders, we align their representations from two complementary perspectives: spatial distribution and scene semantic structure.
As shown in Figure~\ref{fig:pipeline}, Spatial Dependency Alignment (SDA) aligns the spatial attention distribution of the planner representation with that of the VFM representation, encouraging the planner to capture fine-grained spatial dependencies.
Scene Semantic Alignment (SSA) further aligns the global latent representations of the E2E planner and the VFMs through a bidirectional contrastive objective, preserving scene-level semantic consistency across training samples.
Together, these two objectives improve the representation quality of the E2E planner encoder for downstream trajectory planning.

\textbf{Spatial Dependency Alignment.}
SDA aims to transfer the spatial relational structure encoded by the VFM to the E2E planner representation.
Different from directly matching feature values at each spatial location, we align the spatial dependency distributions between planner and VFM features, following the spirit of attention-based knowledge distillation~\citep{zagoruyko2016paying, tung2019similarity}.
Given projected and flattened planner and VFM features, we define the SDA loss as
\begin{equation}
\mathcal{L}_{\mathrm{SDA}}
=
\mathbb{E}_{b,i}
\left[
D_{\mathrm{KL}}
\left(
\mathbf{A}^{t}_{b,i,:}
\,\Vert\,
\mathbf{A}^{s}_{b,i,:}
\right)
\right],
\label{eq:spatial_alignment_loss}
\end{equation}
where the expectation is taken over all samples in the batch and all spatial tokens.
Here, $D_{\mathrm{KL}}$ denotes the Kullback--Leibler divergence,
$\mathbf{A}^{s}=\operatorname{softmax}(\mathbf{Q}^{s}(\mathbf{K}^{t})^\top/\gamma_{sda})$
and
$\mathbf{A}^{t}=\operatorname{softmax}(\mathbf{Q}^{t}(\mathbf{K}^{t})^\top/\gamma_{sda})$
denote the planner-to-VFM dependency distribution and the VFM self-dependency distribution, respectively. $\mathbf{Q}^{s}$ and $\mathbf{Q}^{t}$ are query projections of the planner and VFM features, and $\mathbf{K}^{t}$ is the key projection of the VFM features. We set the SDA temperature to $\gamma_{\mathrm{sda}} = 0.07$.
This objective encourages the planner encoder to reproduce the spatial dependencies distilled from the frozen VFM.

\textbf{Scene Semantic Alignment.}
SSA encourages the planner representation to preserve global semantic consistency with the VFM, following the standard contrastive learning formulation~\citep{oord2018representation, chen2020simple, radford2021learning}. After aggregating the projected planner and VFM features into global embeddings $\mathbf{z}^{s}$ and $\mathbf{z}^{t}$, we apply a bidirectional contrastive loss:
\begin{equation}
\mathcal{L}_{\mathrm{SSA}}
=
\frac{1}{2}
\left[
\operatorname{CE}(\mathbf{M}, \mathbf{y})
+
\operatorname{CE}(\mathbf{M}^{\top}, \mathbf{y})
\right],
\label{eq:semantic_alignment_loss}
\end{equation}
where $\operatorname{CE}$ denotes the cross-entropy loss,
$\mathbf{M}=\mathbf{z}^{s}(\mathbf{z}^{t})^\top/\gamma_{\mathrm{ssa}}$
is the planner--VFM similarity matrix, 
$\mathbf{y}=[1,\dots,B]$
denotes the matched scene indices within the batch, with \(B\) being the batch size. We set the SSA temperature to $\gamma_{\mathrm{ssa}} = 1.0$.
The row-wise and column-wise cross-entropy terms enforce planner-to-VFM and VFM-to-planner matching, respectively.

\subsection{Training Objective}
\label{sec:objective}
Different E2E planners use different training objectives depending on their architectural designs.
Some are trained with trajectory supervision only, while others include auxiliary perception objectives, such as detection, segmentation, or depth estimation.
ViRA is compatible with both settings: we keep the original E2E planner architecture and training protocol unchanged, and simply add representation alignment losses as representation-level supervision.
Let $\mathcal{L}_{\mathrm{planner}}$ denote the original training objective of a planner, including its trajectory loss and any native auxiliary supervision if used.
The overall training objective is:
\begin{equation}
\mathcal{L}
=
\mathcal{L}_{\mathrm{planner}}
+
\lambda_{\mathrm{SDA}}\mathcal{L}_{\mathrm{SDA}}
+
\lambda_{\mathrm{SSA}}\mathcal{L}_{\mathrm{SSA}} .
\label{eq:total_loss_appendix}
\end{equation}
Here, $\mathcal{L}_{\mathrm{SDA}}$ and $\mathcal{L}_{\mathrm{SSA}}$ denote the proposed spatial and semantic latent alignment losses, respectively.
Motivated by recent findings that spatial structure is more critical than global semantics in representation alignment~\citep{singh2025irepa}, we set $\lambda_{\mathrm{SDA}}=1$ and $\lambda_{\mathrm{SSA}}=0.1$.
For planning-only planners, $\mathcal{L}_{\mathrm{planner}}$ contains only the trajectory prediction objective.
For planners with auxiliary objectives, the original auxiliary supervision terms are included in $\mathcal{L}_{\mathrm{planner}}$ and kept unchanged.

\begin{table}[!htbp]
\centering
\caption{\textbf{Model and training hyperparameters.}}
\label{s-tab:training_details}
\small
\setlength{\tabcolsep}{2.0pt}
\renewcommand{\arraystretch}{1.05}
\resizebox{\textwidth}{!}{

\begin{tabular}{l|ccc}

\toprule
\multirowcell{1}[0ex][c]{\textbf{Hyperpara.}}

& \multirowcell{1}[0ex][c]{\textbf{TransFuser}~\citep{Kashyap2023Transfuser}}
& \multirowcell{1}[0ex][c]{\textbf{DiffusionDrive}~\citep{liao2025diffusiondrive}}
& \multirowcell{1}[0ex][c]{\textbf{Rap$^\ast$}~\citep{feng2025rap3drasterizationaugmented}} \\

\hline

\multicolumn{4}{c}{\cellcolor[HTML]{E8E8E8}\textit{Model Configuration}} \\

Sensors

&$3 \times \text{Cam.}$
&$3 \times \text{Cam.}$
&$4 \times \text{Cam.}$

\\

Resolution

&$2048 \times 512$
&$2048 \times 512$
&$4 \times 768 \times 448$
\\

Horizon
&$4 \mathrm s$
&$4 \mathrm s$
&$5 \mathrm s$
\\

Frequency
&$2 \mathrm{Hz}$
&$2 \mathrm{Hz}$
&$2 \mathrm{Hz}$
\\

Backbone
& R34~\citep{he2016deep}
& R34~\citep{he2016deep}
& R34~\citep{he2016deep}
\\

Parameters
& $56\mathrm M$
& $61\mathrm M$
& $69\mathrm M$
\\

Aux. Tasks
& Det. Seg.
& Det. Seg.
& None
\\

\hline
\multicolumn{4}{c}{\cellcolor[HTML]{E8E8E8}\textit{Training Configuration}} \\
GPUs
& $8 \times \text{H100}$ 
& $8 \times \text{H100}$ 
& $8\times \text{H100}$ 
\\

Epochs
& $100$
& $100$
& $30$
\\

Total BS
& $192$
& $192$
& $48$
\\

Initial LR
& $1\times{10}^{-4}$
& $6\times{10}^{-4}$
& $1\times{10}^{-4}$
\\

Schedule
& Constant
& Cosine Decay
& Cosine Decay
\\

Optimizer
& Adam~\citep{kingma2014adam}
& AdamW~\citep{loshchilov2017adamw}
& Adam~\citep{kingma2014adam}
\\

\bottomrule
\end{tabular}
}

\end{table}
\section{Implementation and Evaluation Details}
\label{app:impl}
\subsection{Benchmark and Metrics}
\paragraph{NAVSIM.}
We evaluate driving performance on navtest and navhard using NAVSIM v2~\citep{cao2025navsimv2}.
The navtest evaluation covers 12,146 real-world scenarios in a single stage, providing a broad assessment across driving conditions.
For navhard, we follow the same two-stage evaluation setup as SimScale~\citep{tian2025simscale}, using 244 challenging real-world scenarios and 4,164 synthetic follow-up scenarios rendered with 3D Gaussian Splatting.
Both evaluations use the Extended Predictive Driver Model Score (EPDMS)~\citep{li2025hydramdpplusplus,cao2025navsimv2}, which combines multiplicative penalties with a weighted average of driving-quality scores.
\begin{equation}
\mathrm{EPDMS}
=
\underbrace{
    \prod_{m\in\mathcal{M}_{\mathrm{pen}}} S_m
}_{\text{penalties}}
\cdot
\underbrace{
    \frac{
        \sum_{m\in\mathcal{M}_{\mathrm{avg}}} w_m S_m
    }{
        \sum_{m\in\mathcal{M}_{\mathrm{avg}}} w_m
    }
}_{\text{weighted average}},
\end{equation}
where $S_m$ denotes the evaluated subscore for metric $m$, including human-reference filtering where enabled, and $w_m$ is its weight.
The penalty set $\mathcal{M}_{\mathrm{pen}}$ comprises No-at-fault Collisions (NC), Drivable Area Compliance (DAC), Driving Direction Compliance (DDC), and Traffic Light Compliance (TLC).
The weighted set $\mathcal{M}_{\mathrm{avg}}$ comprises Time-to-Collision (TTC), Ego Progress (EP), Lane Keeping (LK), History Comfort (HC), and Extended Comfort (EC).
Human-reference filtering assigns a subscore of one when the corresponding human-reference score is zero.
For navhard, evaluation uses reactive traffic agents and combines scores from the initial and synthetic follow-up scenarios through two-stage aggregation.

\paragraph{HUGSIM.}
HUGSIM~\citep{zhou2025hugsim} is a real-time, photorealistic closed-loop simulator that reconstructs 3D scenes from captured RGB images with 3D Gaussian Splatting and updates ego and actor states at each step from the planner output, so that surrounding vehicles react to the ego.
We evaluate planners trained only on NAVSIM in a zero-shot manner on scenarios grouped into four difficulty levels: Easy, Medium, Hard, and Extreme.
Each episode is scored by HD-Score~\citep{zhou2025hugsim}, reported on a $0$--$100$ scale per level and on average:
\begin{equation}
\label{eq:hdscore}
    \text{HD-Score} = \text{RC} \times \frac{1}{T}\sum_{t=1}^{T}
    \left(
        \text{NC}_{t} \times \text{DAC}_{t} \times
        \frac{5\cdot\text{TTC}_{t} + 2\cdot\text{Comf}_{t}}{7}
    \right),
\end{equation}
where RC is the fraction of the intended route completed within an episode of $T$ steps, NC and DAC are defined as above, and TTC and Comfort (Comf.) take values in $[0,1]$.

\paragraph{mIoU.}
We measure the semantics encoded in planner representations with the BEV segmentation head of TransFuser.
Mean Intersection-over-Union (mIoU) is the unweighted average of per-class IoU over seven classes: background, road, walkways, centerline, static objects, vehicles, and pedestrians.
\begin{equation}
    \mathrm{IoU}_{c}
    =
    \frac{|P_{c} \cap G_{c}|}{|P_{c} \cup G_{c}|},
    \quad
    \mathrm{mIoU}
    =
    \frac{1}{C}\sum_{c=1}^{C}\mathrm{IoU}_{c},
    \end{equation}
    where $C=7$, $P_{c}$ is the set of BEV cells predicted as class $c$ over the evaluation set, and $G_{c}$ is the corresponding ground-truth set.
The main-text table reports class-wise IoU only for background, road, walkways, centerline, and vehicles.
Static objects and pedestrians remain below 0.5 IoU for every evaluated target, so they are omitted from that table; the reported mIoU still averages all seven classes.
Table~\ref{tab:bev_miou} lists the score of every class.
\begin{table}[!htbp]
\centering
\caption{\textbf{
Effect of alignment objectives on navtest.}
MSE, Cosine, KL, and Contrastive denote mean squared error, cosine similarity, Kullback--Leibler divergence, and contrastive loss, respectively.}
\label{tab:table3_ablation}
\small
\setlength{\tabcolsep}{2pt}
\begin{tabular}{cccc|ccccc}
\toprule
\textbf{MSE}
& \textbf{Cosine}
& \textbf{KL}
& \textbf{Contrastive}
& \textbf{DAC}
& \textbf{DDC}
& \textbf{TLC}
& \textbf{LK}
& \textbf{EPDMS} $\uparrow$ \\

\cmark
& \xmark
& \xmark
& \xmark
& 90.9
& 98.6
& 99.8
& 92.3
& 84.4 \\

\xmark
& \cmark
& \xmark
& \xmark
& 90.8
& 99.0
& 99.8
& 95.6
& 85.1 \\

\xmark
& \xmark
& \cmark
& \xmark
& 90.3
& 98.9
& 99.7
& 94.6
& 84.3 \\

\xmark
& \xmark
& \xmark
& \cmark
& 90.4
& 98.9
& 99.8
& 95.8
& 84.6 \\

\midrule
\cmark
& \cmark
& \xmark
& \xmark
& 89.6
& 99.0
& 99.7
& 95.9
& 83.6 \\

\cmark
& \xmark
& \cmark
& \xmark
& 92.9
& 99.4
& 99.8
& 95.8
& 87.8 \\

\xmark
& \cmark
& \xmark
& \cmark
& 93.5
& 99.4
& 99.8
& 96.4
& 88.1 \\

\xmark
& \xmark
& \cmark
& \cmark
& 94.1
& 99.5
& 99.8
& 96.8
& \textbf{89.0} \\

\bottomrule
\end{tabular}
\end{table}
\subsection{Training Details}
This section complements the experimental setup in Section~\ref{sec:exp_setup} by providing the model configurations and training hyperparameters for all evaluated planners.
As summarized in Table~\ref{s-tab:training_details}, we follow the NAVSIM~\citep{dauner2024navsim} training protocol unless otherwise specified, and train all models on the training logs~\citep{karnchanachari2024towards} of navtrain.
Each planner is trained with its corresponding official settings, while ViRA leaves the planner architecture and inference pipeline unchanged.

\paragraph{TransFuser}~\citep{Kashyap2023Transfuser} is a regression-based planner built on fused sensory representations.
It uses three camera views with a resolution of $2048 \times 512$, predicts a $4\mathrm{s}$ future trajectory at $2\mathrm{Hz}$, and adopts ResNet-34~\citep{he2016deep} as the image backbone.
The model contains $56\mathrm{M}$ parameters and uses detection and segmentation as auxiliary tasks.
We train TransFuser for $100$ epochs on $8$ H100 GPUs with a total batch size of $192$, an initial learning rate of $1\times10^{-4}$, a constant learning-rate schedule, and the Adam optimizer~\citep{kingma2014adam}.

\paragraph{DiffusionDrive}~\citep{liao2025diffusiondrive} is a generative planner that predicts future trajectories through diffusion-based trajectory modeling.
It uses the same three-camera input resolution, prediction horizon, and frequency as TransFuser, i.e., $2048 \times 512$ images, a $4\mathrm{s}$ horizon, and $2\mathrm{Hz}$ trajectory frequency.
It also adopts ResNet-34~\citep{he2016deep} as the image backbone and uses detection and segmentation auxiliary tasks.
The model contains $61\mathrm{M}$ parameters.
We train DiffusionDrive for $100$ epochs on $8$ H100 GPUs with a total batch size of $192$, an initial learning rate of $6\times10^{-4}$, a cosine decay schedule, and the AdamW optimizer~\citep{loshchilov2017adamw}.

\paragraph{Rap$^\ast$}~\citep{feng2025rap3drasterizationaugmented} is a scoring-based planner that selects trajectories from a predefined trajectory vocabulary.
It takes four camera views with a resolution of $4 \times 768 \times 448$, predicts a $5\mathrm{s}$ future trajectory at $2\mathrm{Hz}$, and uses ResNet-34~\citep{he2016deep} as the image backbone.
The corresponding model size is $69\mathrm{M}$.
Unlike TransFuser and DiffusionDrive, Rap$^\ast$ does not use auxiliary detection or segmentation tasks.
We train Rap$^\ast$ for $30$ epochs on $8$ H100 GPUs with a total batch size of $48$, an initial learning rate of $1\times10^{-4}$, a cosine decay schedule, and the Adam optimizer~\citep{kingma2014adam}.

\section{Additional Quantitative Experiments}
\label{app:add_exp}
\subsection{Latent Alignment Implementation}
\label{app:la_details}

We provide additional implementation details of the representation alignment module.
In our experiments, we use the official public checkpoints of DVGT~\citep{zuo2025dvgt}, VGGT~\citep{wang2025vggt}, DINOv3~\citep{simeoni2025dinov3}, SAM3~\citep{carion2025sam}, and DA3~\citep{lin2025da3} as frozen VFM targets, and keep all VFM parameters frozen throughout training.
For all VFM targets, we use the intermediate representations before their task-specific prediction heads, rather than the final task outputs.
This design avoids relying on explicit pseudo-labels and instead transfers the VFM's learned visual representations to the planner encoder.

For DVGT~\citep{zuo2025dvgt}, we follow its original feature extraction design and take the multi-level features before the DPT prediction head.
Specifically, we extract four intermediate token features from the DVGT aggregator, fuse them with learnable weights, and reshape the patch tokens into a spatial feature map.
For other VFM targets, we follow the same principle and use the corresponding feature maps before their prediction heads.

Given the planner feature $\mathbf{F}^{s}$ and the VFM feature $\mathbf{F}^{t}$, we first resize $\mathbf{F}^{s}$ to the spatial resolution of $\mathbf{F}^{t}$.
Both features are then projected into a shared latent space by lightweight $1{\times}1$ convolutional projection heads.
The projected features are used to compute the spatial distribution alignment loss and the semantic consistency loss described in Appendix~\ref{sec:LA}.
The VFM and alignment heads are used only during training and discarded after training, introducing no additional inference cost.

\subsection{Different Alignment Objectives Ablation}
To validate the effectiveness of the latent alignment objective, we conduct an ablation study on TransFuser~\citep{Kashyap2023Transfuser} with different alignment objectives.
We consider representative losses for feature-level, distribution-level, and contrastive supervision, and compare different combinations for spatial and semantic alignment~\citep{yu2024repa,li2025sf,Wu2026gf}.
As shown in Table~\ref{tab:table3_ablation}, jointly aligning spatial dependencies and semantic consistency generally yields better planning performance than using a single alignment objective.
Among all combinations, KL divergence for spatial alignment and contrastive loss for semantic alignment achieve the best performance, reaching an EPDMS of 89.0.
\begin{figure}[!htbp]
  \centering
  \includegraphics[width=1.0\textwidth]{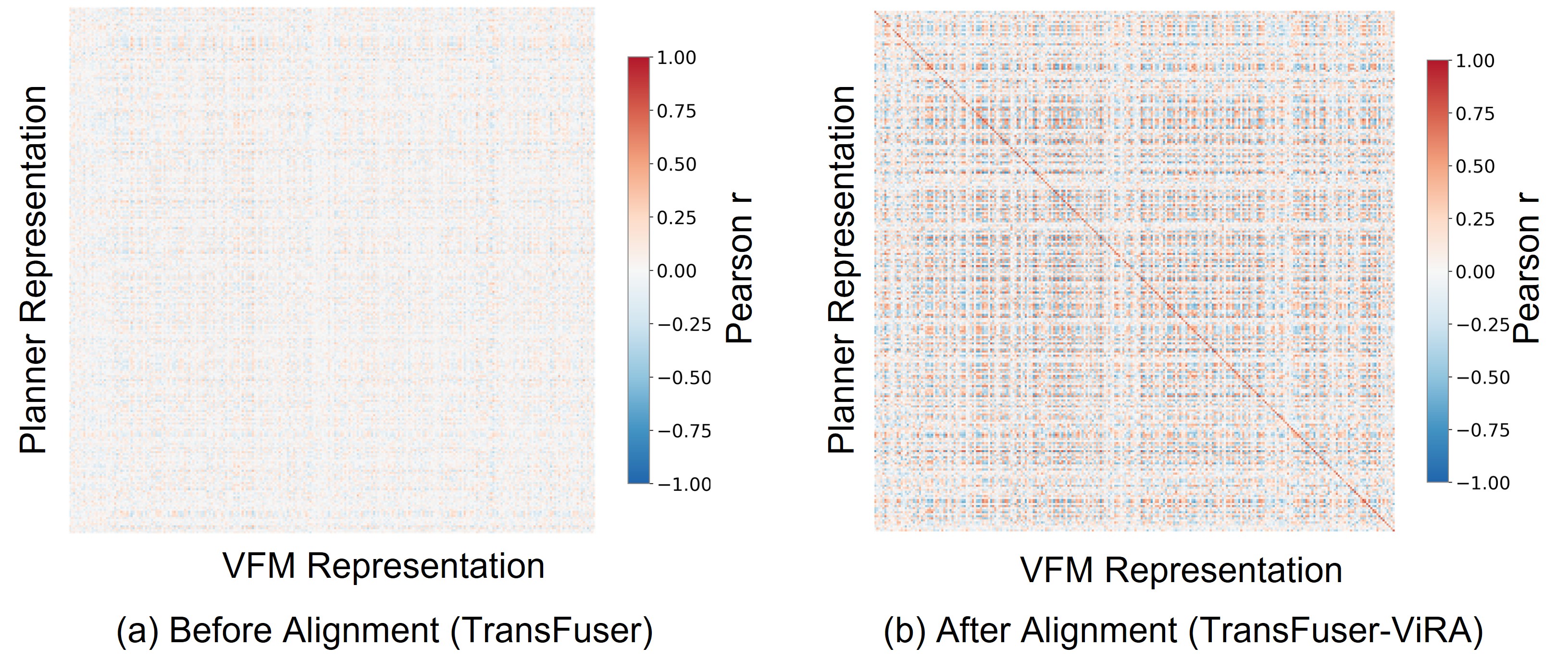}
  \caption{\textbf{Feature correlation analysis.} }
  \label{fig:correlation}
\end{figure}
\subsection{Supplementary Alignment Analysis}
\label{app:correlation}
To further verify the alignment effectiveness of ViRA, we compute the Pearson correlation between planner encoder features and VFM features.
As shown in Figure~\ref{fig:correlation}, the original TransFuser encoder exhibits weak correlation with the VFM features, and the resulting correlation matrix lacks clear structural patterns.
After DVGT supervision, TransFuser with DVGT shows stronger and more structured correlation patterns, indicating that ViRA effectively narrows the representation gap between the planner encoder and the VFM.
This result provides feature-level evidence for the effectiveness of ViRA.

\subsection{VFM Target Dependency Analysis}
\label{app:teacher_analysis}
This section provides additional quantitative results for the VFM-target analysis.
We evaluate ViRA on both TransFuser~\citep{Kashyap2023Transfuser} and Rap$^\ast$~\citep{feng2025rap3drasterizationaugmented}.
TransFuser uses detection and segmentation as perception auxiliary tasks, whereas Rap$^\ast$ does not rely on explicit perception supervision, allowing us to examine whether ViRA behaves consistently across different planner designs.
\begin{table}[!ht]
\caption{\textbf{Effect of different VFM teachers on TransFuser.} Performance on the NAVSIM v2 navtest benchmark.}
\label{tab:lft_with_dif_teacher}
\small
\setlength{\tabcolsep}{2.0pt}
\renewcommand{\arraystretch}{1.05}
\resizebox{\linewidth}{!}{
\begin{tabular}{l|llllllllll}
\toprule 
{\textbf{Teachers}}
& {\textbf{NC} $\uparrow$} 
& {\textbf{DAC} $\uparrow$} 
& {\textbf{DDC} $\uparrow$} 
& {\textbf{TLC} $\uparrow$} 
& {\textbf{EP} $\uparrow$} 
& {\textbf{TTC} $\uparrow$} 
& {\textbf{LK} $\uparrow$} 
& {\textbf{HC} $\uparrow$} 
& {\textbf{EC} $\uparrow$} 
& \textbf{EPDMS} $\uparrow$ 
\\ 

\midrule
{Baseline}
& {97.1} 
& {89.6} 
& {99.0} 
& {99.8} 
& {97.9} 
& {95.9} 
& {95.9} 
& {98.3} 
& {87.3} 
& {83.6} 
\\ 

\midrule
{Random init}
& {96.8}
& {90.2}
& {99.0}
& {99.7}
& {98.0}
& {96.1}
& {95.5}
& {98.3}
& {86.7}
& 82.0 \textcolor[HTML]{0073CF}{{\footnotesize -1.6}}

\\

{DVGT~\citep{zuo2025dvgt}}
& {97.7} 
& {94.1} 
& {99.5} 
& {99.8} 
& {98.2} 
& {96.7} 
& {96.8} 
& {98.3} 
& {87.9} 
& {89.0 \textcolor[HTML]{C71585}{{\footnotesize +5.4}}} 
\\

{VGGT~\citep{wang2025vggt}}
& {97.8} 
& {94.6} 
& {99.3} 
& {99.8} 
& {97.8} 
& {97.0} 
& {96.7} 
& {98.3} 
& {87.2} 
& {89.3 \textcolor[HTML]{C71585}{{\footnotesize +5.7}}} 
\\

{DA3~\citep{lin2025da3}}
& {97.9} 
& {94.2} 
& {99.5} 
& {99.8} 
& {98.0} 
& {97.0} 
& {95.9} 
& {98.3} 
& {87.5} 
& {89.1 \textcolor[HTML]{C71585}{{\footnotesize +5.5}}} 
\\

{DINOv3-L~\citep{simeoni2025dinov3}}
& {97.6} 
& {94.4} 
& {99.3} 
& {99.7} 
& {98.0} 
& {97.0} 
& {96.7} 
& {98.3} 
& {86.6} 
& {88.8 \textcolor[HTML]{C71585}{{\footnotesize +5.2}}} 
\\

{SAM3~\citep{carion2025sam}}
& {97.9} 
& {94.2} 
& {99.5} 
& {99.8} 
& {98.0} 
& {96.9} 
& {95.1} 
& {98.3} 
& {88.1} 
& {89.0 \textcolor[HTML]{C71585}{{\footnotesize +5.4}}} 
\\

\bottomrule
\end{tabular}}
\end{table}
\begin{table}[!ht]
\caption{\textbf{Effect of different VFM teachers on TransFuser without auxiliary perception supervision.} Performance on the NAVSIM v2 navtest benchmark.}
\label{tab:lft_without_per_dif_teacher}
\small
\setlength{\tabcolsep}{2.0pt}
\renewcommand{\arraystretch}{1.05}
\resizebox{\linewidth}{!}{
\begin{tabular}{l|llllllllll}
\toprule 
{\textbf{Teachers}}
& {\textbf{NC} $\uparrow$} 
& {\textbf{DAC} $\uparrow$} 
& {\textbf{DDC} $\uparrow$} 
& {\textbf{TLC} $\uparrow$} 
& {\textbf{EP} $\uparrow$} 
& {\textbf{TTC} $\uparrow$} 
& {\textbf{LK} $\uparrow$} 
& {\textbf{HC} $\uparrow$} 
& {\textbf{EC} $\uparrow$} 
& \textbf{EPDMS} $\uparrow$ 
\\ 

\midrule
{Baseline}
& {97.0} 
& {89.9} 
& {98.5} 
& {99.8} 
& {97.9} 
& {96.3} 
& {94.3} 
& {98.4} 
& {86.8} 
& {82.6} 
\\ 

\midrule
{Random init}
& {95.4}
& {90.9}
& {97.7}
& {99.5}
& {98.1}
& {94.2}
& {93.6}
& {98.3}
& {84.8}
& 80.1 \textcolor[HTML]{0073CF}{{\footnotesize -2.5}}

\\

{DVGT~\citep{zuo2025dvgt}}
& {97.9} 
& {90.9} 
& {99.1} 
& {99.7} 
& {98.0} 
& {96.8} 
& {96.5} 
& {98.3} 
& {87.2} 
& {88.2 \textcolor[HTML]{C71585}{{\footnotesize +5.6}}} 
\\

{VGGT~\citep{wang2025vggt}}
& {97.7} 
& {92.9} 
& {99.1} 
& {99.8} 
& {98.0} 
& {96.6} 
& {95.9} 
& {98.3} 
& {87.5} 
& {87.6 \textcolor[HTML]{C71585}{{\footnotesize +5.0}}} 
\\

{DA3~\citep{lin2025da3}}
& {97.9} 
& {92.4} 
& {99.1} 
& {99.7} 
& {97.9} 
& {96.7} 
& {95.9} 
& {98.3} 
& {85.8} 
& {87.2 \textcolor[HTML]{C71585}{{\footnotesize +4.6}}} 
\\

{DINOv3-L~\citep{simeoni2025dinov3}}
& {97.9} 
& {90.9} 
& {99.3} 
& {99.8} 
& {98.2} 
& {96.9} 
& {96.7} 
& {98.5} 
& {88.6} 
& {89.1 \textcolor[HTML]{C71585}{{\footnotesize +6.5}}} 
\\

{SAM3~\citep{carion2025sam}}
& {97.6} 
& {91.5} 
& {99.0} 
& {99.8} 
& {97.9} 
& {96.4} 
& {96.0} 
& {98.4} 
& {88.2} 
& {86.4 \textcolor[HTML]{C71585}{{\footnotesize +3.8}}} 
\\

\bottomrule
\end{tabular}}
\end{table}
\paragraph{VFM-target dependency across planners.}
Across the five targets used in the main comparison, ViRA improves TransFuser by 5.2–5.7 EPDMS and Rap* by 6.6–12.1 EPDMS (Tables~\ref{tab:lft_with_dif_teacher} and~\ref{tab:rap_with_dif_teacher}).
Table~\ref{tab:lft_without_per_dif_teacher} further shows that ViRA improves TransFuser without auxiliary perception supervision by 3.8--6.5 EPDMS points across the five VFM targets.
Compared with Table~\ref{tab:lft_with_dif_teacher}, removing auxiliary perception supervision increases the EPDMS spread across the same targets from 0.5 to 2.7 points, indicating greater sensitivity to VFM target selection.

\paragraph{Random-target control.}
To test whether the gains depend on VFM pre-training, we replace the pre-trained target with a randomly initialized VFM target while retaining the same alignment objectives.
As shown in Tables~\ref{tab:lft_with_dif_teacher} and~\ref{tab:rap_with_dif_teacher}, random-target alignment lowers EPDMS by $1.6$ points for TransFuser and $1.7$ points for Rap$^\ast$.
Pre-trained DVGT improves both planners, by $+5.4$ and $+10.9$ points, respectively.
Applying the same alignment objectives to random features therefore does not reproduce the gains observed with pre-trained VFM representations.

\paragraph{BEV semantic map analysis.}
We further report the BEV semantic map results of TransFuser in Table~\ref{tab:bev_miou}.
All pre-trained VFM targets improve BEV segmentation mIoU, but the gains vary from $+0.5$ (SAM3) to $+3.0$ (DINOv3).
\section{Additional Qualitative Visualizations}
\label{app:qual}

\subsection{Additional Qualitative Analysis}
\begin{figure}[!htbp]
    \vspace{15pt}
    \centering                      
    \includegraphics[width=1.0\textwidth]{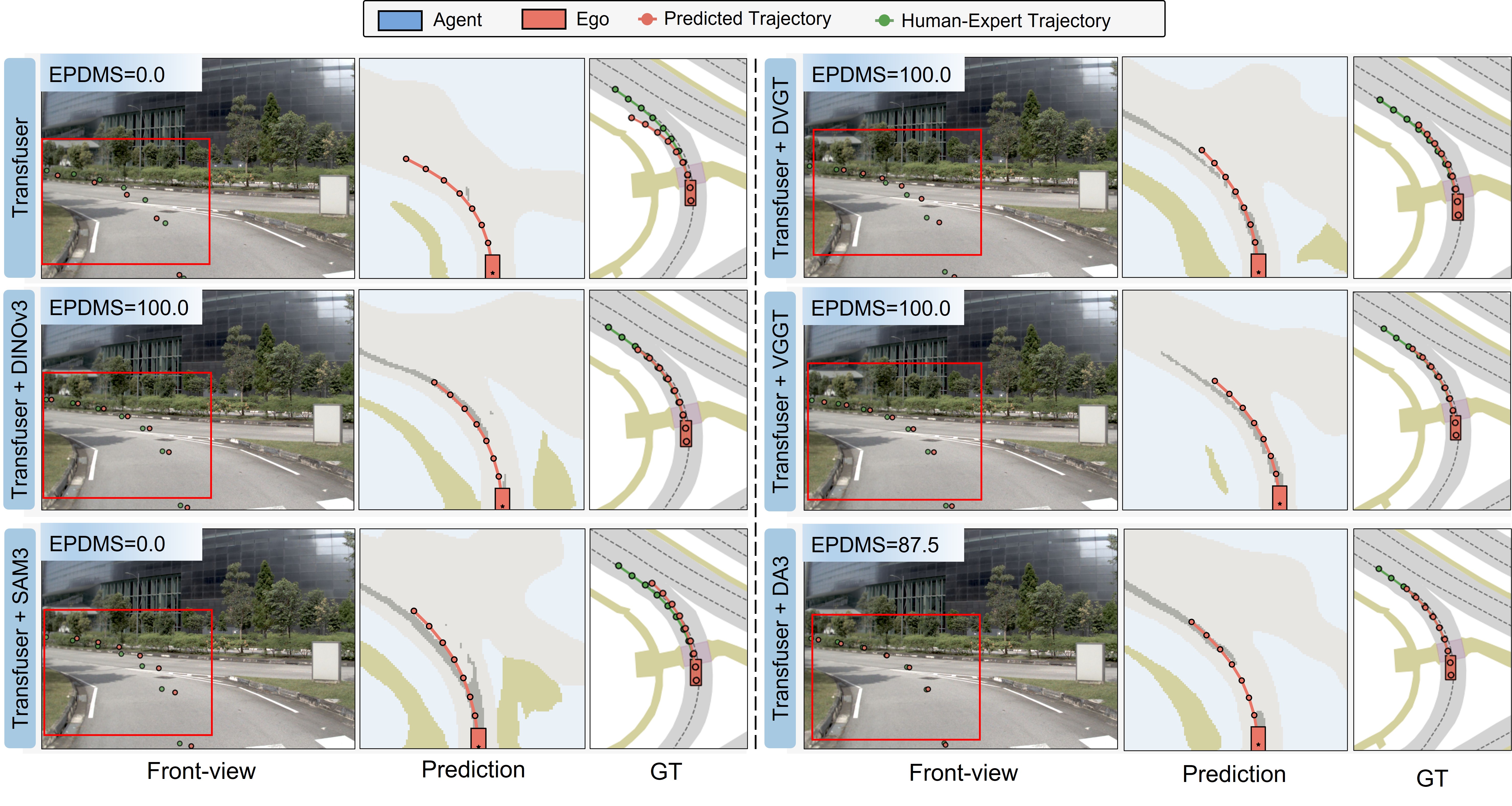}
    \caption{\textbf{Additional qualitative analysis for perception-based planners across different VFMs.}}
    \label{fig:s_aux_dif_teacher}
\end{figure}

\begin{figure}[!htbp]
    \centering
    \includegraphics[width=1.0\textwidth]{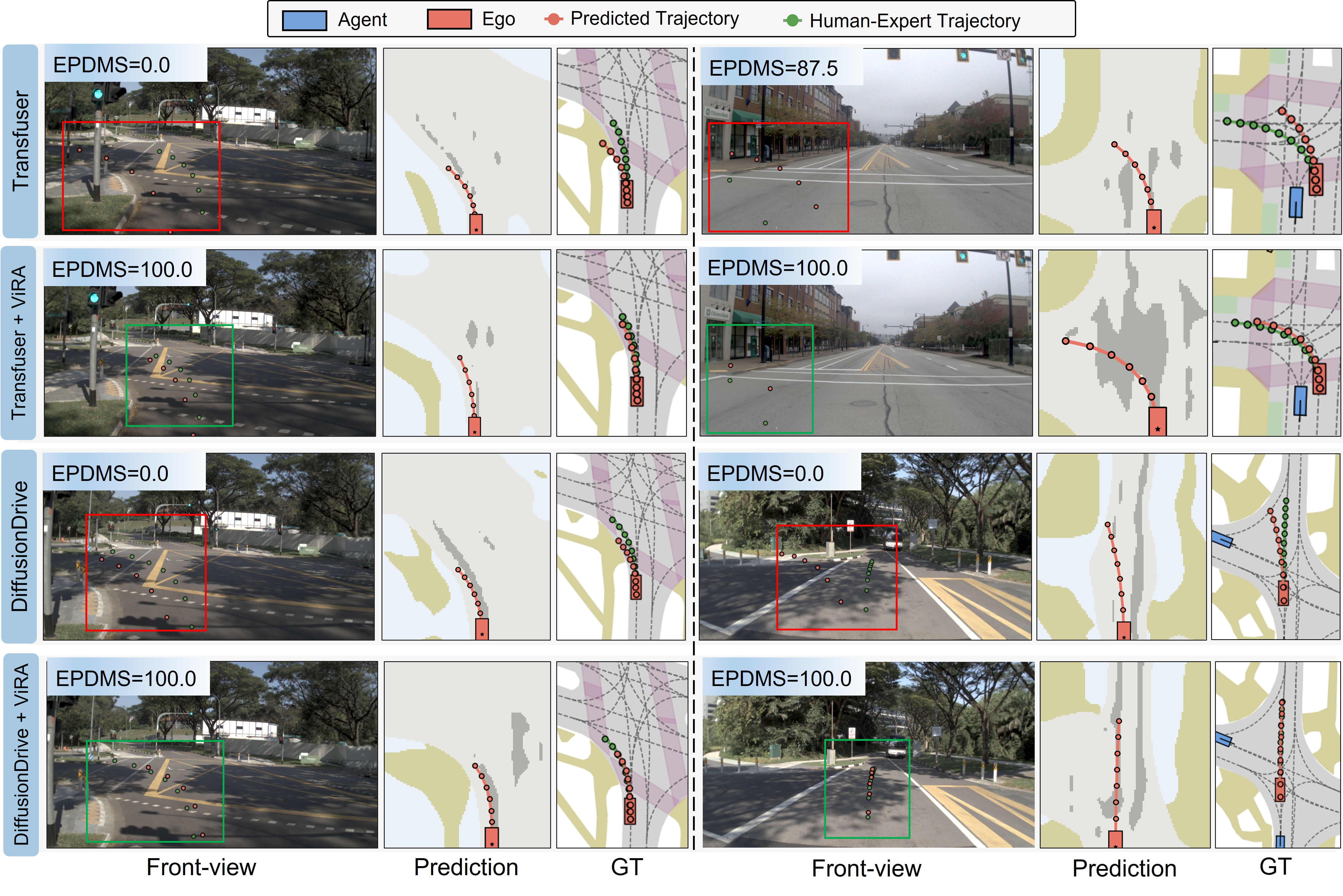}
    \caption{\textbf{Additional qualitative analysis for perception-based planners.}}
    \label{fig:s_aux_results}
\end{figure}

\begin{figure}[!htbp]
    \centering                      
    \includegraphics[width=1.0\textwidth]{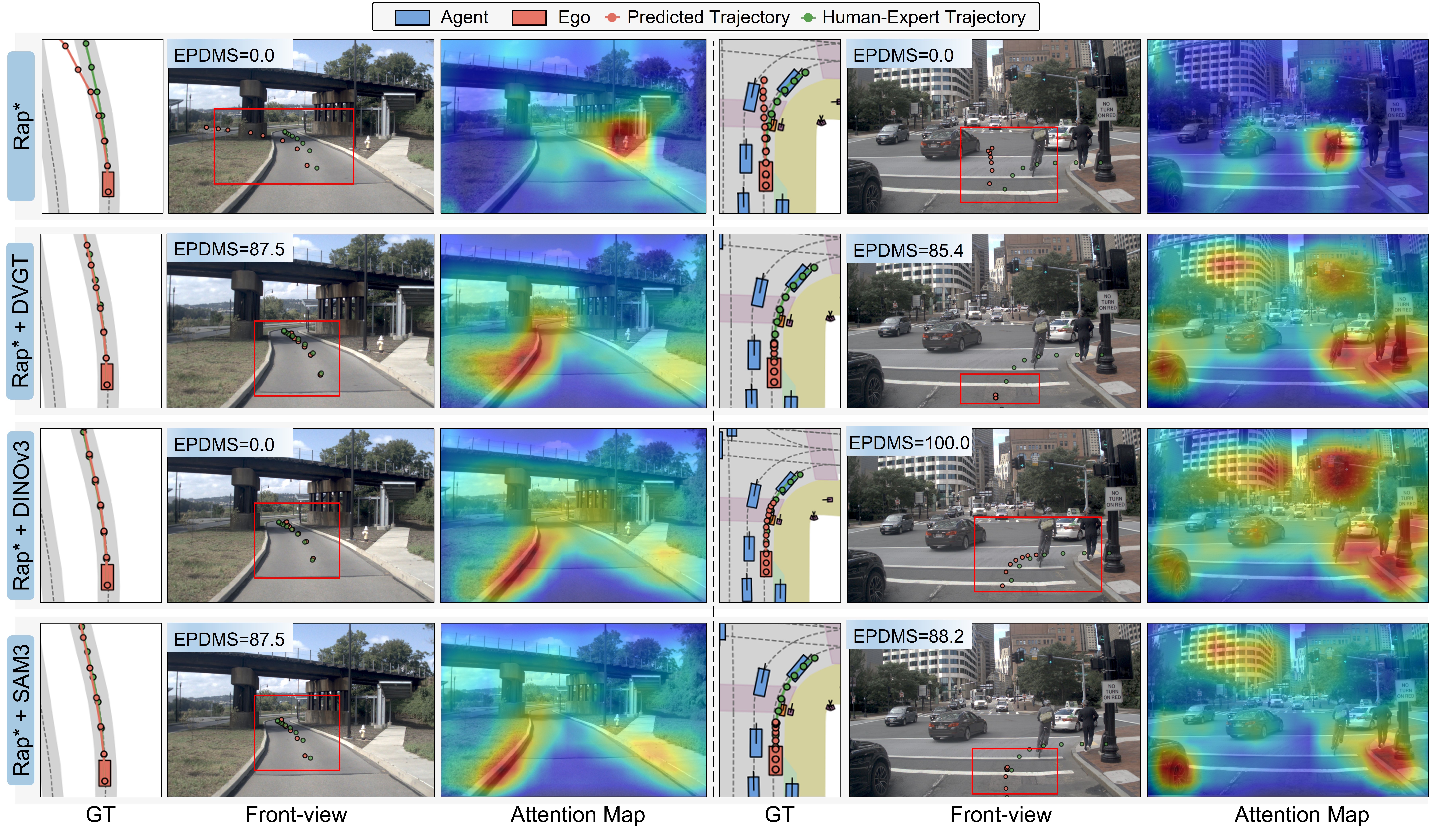}
    \caption{\textbf{Additional qualitative analysis for perception-free planners across different VFMs.}}
    \label{fig:rap_dif_teacher}
\end{figure}

We provide additional qualitative visualizations to further analyze the effect of ViRA across different planners, VFM targets, and driving scenarios.
These examples include perception-based planners in complex intersection scenes Figure~\ref{fig:s_aux_results}, and comparisons under different VFM targets Figure~\ref{fig:s_aux_dif_teacher}. Figure~\ref{fig:rap_dif_teacher} also shows that the benefits of VFM-guided alignment for perception-free planning vary across targets and scenarios.
Together, these visualizations complement the quantitative results and show that ViRA consistently improves scene understanding and planning behavior across different settings.

\subsection{Failure Cases}
\begin{figure}[htbp]
  \centering
  \includegraphics[width=1.0\textwidth]{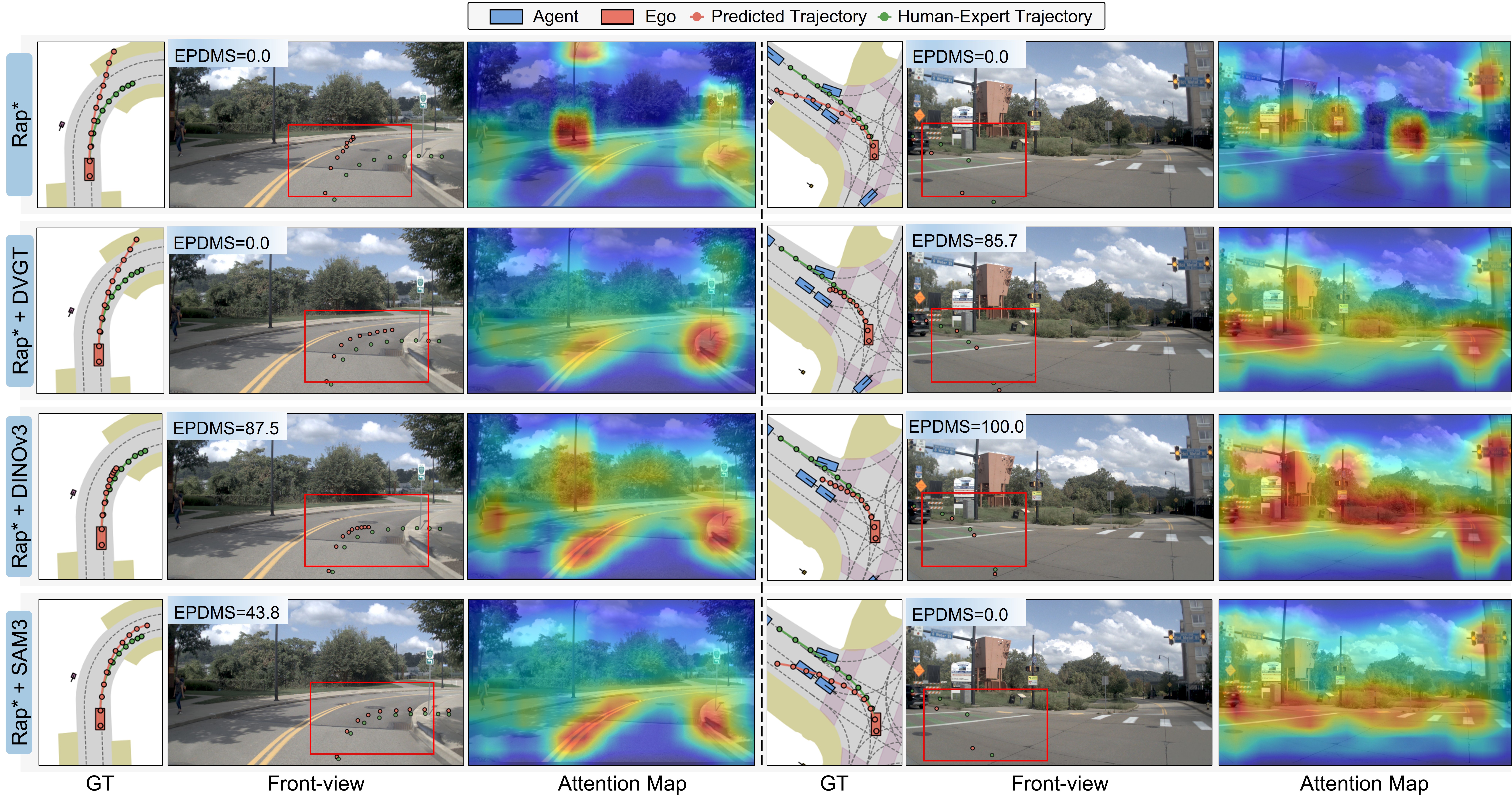}
  \caption{\textbf{Failure case analysis for a perception-free planner.}}
  \label{fig:s_non_aux_failure}
\end{figure}

\begin{figure}[!htbp]
    \centering
    \includegraphics[width=1.0\textwidth]{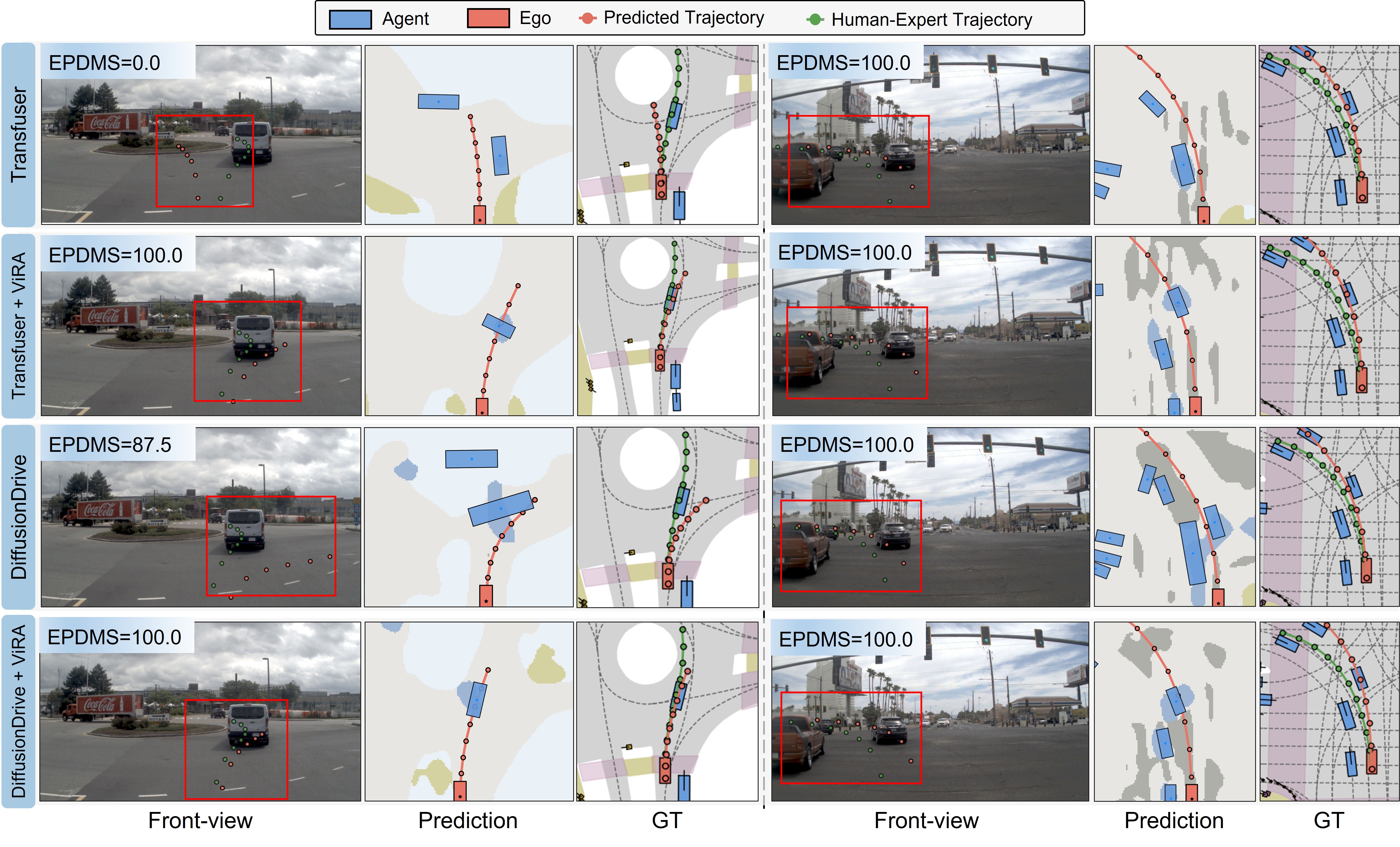}
    \caption{\textbf{Failure case analysis for perception-based planners.}}
    \label{fig:s_aux_failure_case}
\end{figure}

Although ViRA improves representation quality and downstream planning performance, challenging failure cases remain.
For perception-free planners Figure~\ref{fig:s_non_aux_failure}, different VFM targets exhibit slightly different failure patterns, suggesting that their representations may differ in subtle but meaningful ways and lead to small performance variations.
For perception-based planners Figure~\ref{fig:s_aux_failure_case}, even when ViRA produces more complete BEV semantic maps, the planned trajectory may still fail to follow the human expert in challenging scenarios.
These cases show that high-quality visual representations are fundamental, but not a silver bullet for end-to-end autonomous driving.
Robust planning still requires better trajectory decoding, temporal reasoning, interaction modeling, and decision-making under uncertainty.

\begin{table}[!ht]
\caption{\textbf{Effect of different VFM teachers on Rap$^\ast$.} Performance on the NAVSIM v2 navtest benchmark.
Rap$^\ast$ is our reimplementation with a different backbone from the official codebase and without the original data augmentation.
}
\label{tab:rap_with_dif_teacher}
\small
\setlength{\tabcolsep}{2.0pt}
\renewcommand{\arraystretch}{1.05}
\resizebox{\linewidth}{!}{
\begin{tabular}{l|llllllllll} 
\toprule 
{\textbf{Teachers}}
& {\textbf{NC} $\uparrow$} 
& {\textbf{DAC} $\uparrow$} 
& {\textbf{DDC} $\uparrow$} 
& {\textbf{TLC} $\uparrow$} 
& {\textbf{EP} $\uparrow$} 
& {\textbf{TTC} $\uparrow$} 
& {\textbf{LK} $\uparrow$} 
& {\textbf{HC} $\uparrow$} 
& {\textbf{EC} $\uparrow$} 
& \textbf{EPDMS} $\uparrow$ 
\\ 

\midrule
{Baseline}
& {91.6} 
& {88.8} 
& {93.5} 
& {99.8} 
& {87.2} 
& {92.5} 
& {90.9} 
& {97.2} 
& {56.7} 
& {72.8} 
\\ 

\midrule
{Random init}
& 94.6
& 88.1
& 95.1
& 99.8
& 84.2
& 95.1
& 93.1
& 96.7
& 41.9
& 71.1 \textcolor[HTML]{0073CF}{{\footnotesize -1.7}}
\\

{DVGT~\citep{zuo2025dvgt}}
& {97.1} 
& {96.9} 
& {99.2} 
& {99.9} 
& {87.5} 
& {96.9} 
& {96.2} 
& {97.7} 
& {48.1} 
& {83.7 \textcolor[HTML]{C71585}{{\footnotesize +10.9}}} 
\\

{VGGT~\citep{wang2025vggt}}
& {96.4} 
& {95.6} 
& {98.6} 
& {99.9} 
& {84.0} 
& {96.9} 
& {95.1} 
& {97.4} 
& {36.5} 
& {79.4 \textcolor[HTML]{C71585}{{\footnotesize +6.6}}} 
\\

{DA3~\citep{lin2025da3}}
& {96.1}
& {95.6}
& {98.6}
& {99.7}
& {88.8}
& {96.1}
& {96.0}
& {97.7}
& {54.9}
& {82.5 \textcolor[HTML]{C71585}{{\footnotesize +9.7}}}

\\

{DINOv3~\citep{simeoni2025dinov3}}
& {97.6} 
& {96.4} 
& {98.9} 
& {100.0} 
& {87.8} 
& {97.7} 
& {95.9} 
& {98.0} 
& {58.4} 
& {84.9 \textcolor[HTML]{C71585}{{\footnotesize +12.1}}} 
\\

{SAM3~\citep{carion2025sam}}
& {97.1} 
& {95.5} 
& {98.4} 
& {99.8} 
& {89.9} 
& {96.9} 
& {95.5} 
& {96.8} 
& {60.1} 
& {84.0 \textcolor[HTML]{C71585}{{\footnotesize +11.2}}} 
\\

\bottomrule
\end{tabular}}
\end{table}
\begin{table}[!h]
\centering
\caption{\textbf{VFMs dependency on BEV segmentation of TransFuser.} Class-wise BEV segmentation performance of TransFuser under different VFM targets. BG indicates background, SO indicates static objects.}
\label{tab:bev_miou}
\setlength{\tabcolsep}{1.5pt}
\renewcommand{\arraystretch}{1.05}
\resizebox{\linewidth}{!}{
\begin{tabular}{l|cccccccl}
\toprule
\textbf{Teachers}
& \textbf{BG $\uparrow$}
& \textbf{Road $\uparrow$} 
& \textbf{Walkways $\uparrow$}
& \textbf{Centerline $\uparrow$}
& \textbf{SO $\uparrow$}
& \textbf{Vehicles $\uparrow$}
& \textbf{Pedestrians $\uparrow$}
& \textbf{mIoU $\uparrow$} \\

\midrule
Baseline
& 85.4
& 62.3
& 56.3
& 23.6
& 0.1
& 27.1
& 0.0
& 36.4 \\

\midrule
DA3~\citep{lin2025da3}
& 86.6
& 63.8
& 59.1
& 28.2
& 0.3
& 32.3
& 0.0
& 38.6 \textcolor[HTML]{C71585}{{\footnotesize +2.2}}

\\

VGGT~\citep{wang2025vggt}
& 86.5
& 63.9
& 58.5
& 27.6
& 0.3
& 33.6
& 0.0
& 38.6 \textcolor[HTML]{C71585}{{\footnotesize +2.2}}\\

SAM3~\citep{carion2025sam}
& 85.8
& 62.4
& 56.7
& 25.0
& 0.1
& 28.0
& 0.0
& 36.9 \textcolor[HTML]{C71585}{{\footnotesize +0.5}}\\

DVGT~\citep{zuo2025dvgt}
& 86.9
& 64.6
& 59.3
& 28.9
& 0.5
& 33.4
& 0.2
& 39.1 \textcolor[HTML]{C71585}{{\footnotesize +2.7}}\\

DINOv3~\citep{simeoni2025dinov3}
& 87.1
& 64.7
& 60.1
& 29.8
& 0.3
& 33.5
& 0.2
& 39.4 \textcolor[HTML]{C71585}{{\footnotesize +3.0}}\\

\bottomrule
\end{tabular}}
\end{table}
\section{Limitations and Broader Impact}
\label{app:limitation}

\paragraph{Training-time VFM Cost.}
Although ViRA introduces no additional inference overhead, it relies on external frozen VFM representations during training.
Extracting and storing VFM representations may introduce extra training cost, especially when using large-scale VFMs or high-resolution multi-view inputs.
A practical future direction is to reduce this cost through feature caching, lightweight VFM distillation, or selective alignment over only the most informative layers and spatial regions.
This would make VFM-guided representation alignment more scalable for larger datasets and more diverse planner architectures.

\paragraph{Scope of Architectures and Benchmarks.}
This work evaluates ViRA on three representative E2E planners, on NAVSIM v2 navtest and navhard, and in zero-shot closed-loop evaluation on HUGSIM.
It does not cover multi-modal sensor fusion, vision-language-action models, or real-vehicle deployment.
Further validation on a wider set of planners and real-world data is needed to assess how far the observed planner-dependent conversion from representation quality to planning gains generalizes.

\paragraph{Broader Impact.}
By improving the visual representations used by E2E planners, ViRA may contribute to safer and more reliable autonomous driving systems.
Meanwhile, the use of large-scale VFMs also raises potential concerns.
VFMs may inherit biases from their pre-training data, including geographical, weather, cultural, and traffic-distribution biases.
If such biases are transferred to autonomous driving planners without careful validation, they may cause degraded performance in underrepresented regions or rare safety-critical scenarios.
Therefore, we believe that VFM-guided planning should be accompanied by rigorous evaluation across diverse environments, transparent failure analysis, and safety-oriented validation before real-world deployment.

\section{License of Assets}
\label{app:license}
All training and evaluation are conducted on publicly available datasets and benchmarks, including nuPlan~\citep{karnchanachari2024towards}, OpenScene~\citep{openscene2023}, and NAVSIM v2~\citep{cao2025navsimv2}.
We use these datasets only for research purposes and follow their official licenses and terms of use.
We adopt several publicly available VFMs as VFM targets, including DVGT~\citep{zuo2025dvgt}, VGGT~\citep{wang2025vggt}, DINOv3~\citep{simeoni2025dinov3}, SAM3~\citep{carion2025sam}, and DA3~\citep{lin2025da3}.
DVGT is released under the Apache License 2.0.
VGGT, DINOv3, and SAM3 are used under their respective official licenses.
For DA3, we follow the license associated with the specific released model checkpoint used in our experiments.
We also build upon publicly available E2E planners, including Rap$^\ast$~\citep{feng2025rap3drasterizationaugmented}, DiffusionDrive~\citep{liao2025diffusiondrive}, and TransFuser~\citep{Kashyap2023Transfuser}.
Rap$^\ast$ is released under the Apache License 2.0, while DiffusionDrive and TransFuser are released under the MIT License.
We do not redistribute third-party datasets, model weights, or checkpoints.
All third-party assets are used in accordance with their corresponding licenses.
Our released code will include proper attribution and license notices for all used assets.

\end{document}